\documentclass[11pt]{article}

\usepackage[final]{acl}

\usepackage{times}
\usepackage{latexsym}
\usepackage[utf8]{inputenc}
\usepackage{microtype}

\usepackage{amsmath,amsfonts,bm}

\def\eqref#1{equation~\ref{#1}}

\def\1{\bm{1}}

\DeclareMathAlphabet{\mathsfit}{\encodingdefault}{\sfdefault}{m}{sl}
\SetMathAlphabet{\mathsfit}{bold}{\encodingdefault}{\sfdefault}{bx}{n}

\usepackage{float}
\usepackage{graphicx}
\usepackage{placeins}
\usepackage{url}
\usepackage{booktabs}
\usepackage{multirow}
\usepackage{tabularx}
\usepackage[table]{xcolor}

\usepackage[normalem]{ulem}
\useunder{\uline}{\ul}{}
\usepackage{caption}
\usepackage{xspace}
\usepackage{colortbl}

\definecolor{varysorange}{RGB}{255,140,0}
\definecolor{darkblue}{RGB}{0,0,139}
\definecolor{maroon}{RGB}{128,0,0}

\hypersetup{
  pdftitle={E2Rank: Unifying Text Embedding and Listwise Reranking for Effective and Efficient Search},
  pdfauthor={Qi Liu, Yanzhao Zhang, Mingxin Li, Dingkun Long, Pengjun Xie, Jiaxin Mao}
}

\newcommand{\model}{$\text{E}^2$\textsc{Rank}\xspace}

\title{\model: Unifying Text Embedding and Listwise Reranking for Effective and Efficient Search}

\author{
  \textbf{Qi Liu}\textsuperscript{1},
  \textbf{Yanzhao Zhang}\textsuperscript{2},
  \textbf{Mingxin Li}\textsuperscript{2},
  \textbf{Dingkun Long}\textsuperscript{2},
  \textbf{Pengjun Xie}\textsuperscript{2},
  \textbf{Jiaxin Mao}\textsuperscript{1}\thanks{Corresponding author}
\\
  \textsuperscript{1}Renmin University of China, Beijing, China \quad
  \textsuperscript{2}Alibaba Group, Hangzhou, China \\
  \texttt{qiliu6777@gmail.com, maojiaxin@gmail.com} \\
  \url{https://Alibaba-NLP.github.io/E2Rank}
}

\begin{document}
\maketitle

\begin{abstract}
Text embedding models deliver competitive retrieval performance with high efficiency, but their ranking fidelity remains limited compared to LLM-based listwise rerankers, which capture fine-grained query-document and document-document interactions at high computational cost.
We propose \textbf{\model} (\textbf{E}fficient \textbf{E}mbedding-based \textbf{Rank}ing), a unified framework that extends a single text embedding model to perform both retrieval and listwise reranking via continued training under a listwise ranking objective.
The key insight is to treat the listwise prompt---constructed from the query and its top-K candidates---as a pseudo-relevance feedback (PRF) query, enabling reranking via cosine similarity against precomputed document embeddings without autoregressive decoding.
Empirically, \model achieves state-of-the-art results on BEIR, competitive performance on the reasoning-intensive BRIGHT benchmark, significantly lower latency than existing LLM-based rerankers, and improved embedding performance on MTEB---all within a single model.
\end{abstract}

\section{Introduction}

Modern information retrieval (IR) systems predominantly adopt a two-stage "retrieve-then-rerank" architecture to balance efficiency and effectiveness~\citep{matveeva2006multistage}. In the first stage, text embedding models offer efficient similarity search abilities by mapping queries and documents into a shared low-dimensional vector space, enabling real-time and web-scale applications~\citep{karpukhin2020dense}. The advent of large language models (LLMs) has further improved the retrieval performance of these embedding models~\citep{zhu2023llm4ir,ma2023repllama, behnamghaderllm2vec}.
In the second stage, reranking models further refine the initial retrieval results by modeling more fine-grained relevance signals.
Recent advances in large language models have further strengthened reranking performance, particularly through listwise reranking methods that jointly consider the query and the full candidate set~\citep{sun2023rankgpt}.
These LLM-based rerankers can capture both query-document and document-document interactions, leading to state-of-the-art ranking performance across various benchmarks~\citep{sun2023rankgpt, pradeep2023rankzephyr}.

\begin{figure*}[t!]
    \centering
    \includegraphics[width=1.0\textwidth]{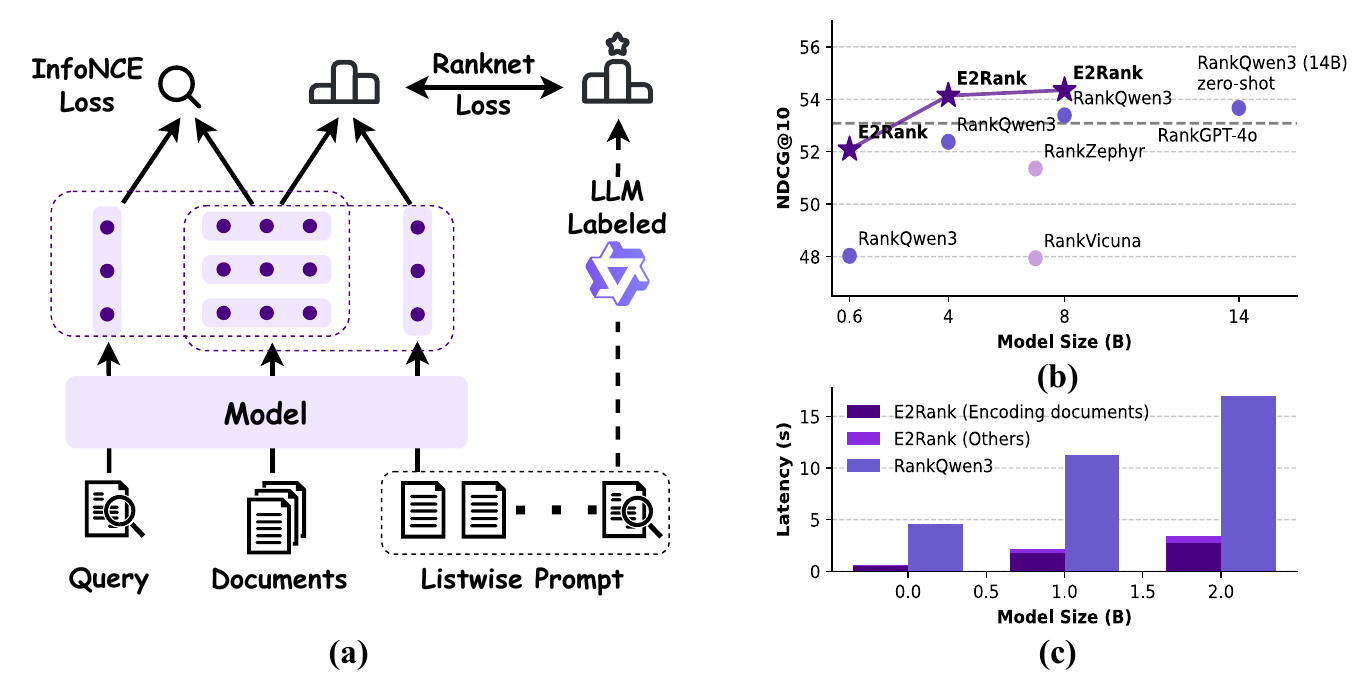}
    \caption{\textbf{(a)} Overview of \model. \textbf{(b)} Average reranking performance on the BEIR benchmark, \model outperforms other baselines. \textbf{(c)} Reranking latency per query on the Covid dataset.}
    \label{fig:cover}
\end{figure*}

Despite their effectiveness, LLM-based listwise rerankers incur high computational costs and inference latency, limiting their deployment in real-time environments. The need to encode all candidates in a single pass introduces substantial prefilling delays, while autoregressive decoding further slows the process~\citep{liu2025perank}.
Therefore, some recent works tried to improve the efficiency of listwise rerankers by compressing the input documents~\citep{liu2025perank} and leveraging LLM's output logits or attention patterns to avoid expensive auto-regressive generation~\citep{reddy2024first, chen2024attention, zhang2025query}.

Among the above works, an important observation is that the auto-regressive generation paradigm adopted by RankGPT~\citep{sun2023rankgpt} is not necessary for ranking, while the interaction between query and documents in the context is critical for ranking effectiveness~\citep{chen2024attention}. Additionally, ~\cite{liu2025perank} shows that incorporating document embeddings in the ranking process is also helpful. 
These observations suggest that the key to effective listwise reranking may lie less in the autoregressive generation process itself and more in the contextual interaction signals encoded in the listwise prompt. This motivates a natural question: \emph{can we transfer such listwise interaction signals into an embedding-based ranking framework, so that reranking can be performed by simple similarity computation rather than autoregressive decoding?}

Intuitively, this question can be addressed from two complementary perspectives. From the standpoint of dense retrieval, the listwise prompt integrating both the document and the query can be viewed as a form of pseudo relevance feedback (PRF)~\citep{xu1996query} query in traditional IR, which can enhance the quality of query embeddings~\citep{yu2021anceprf}. Conversely, from the perspective of listwise reranking, the rich contextual information encoded in the listwise prompt enables the use of simple cosine similarity in place of autoregressive decoding. In essence, the listwise prompt can be transformed into a single PRF-enhanced query embedding, allowing reranking to be efficiently performed via cosine similarity against precomputed document embeddings. This leads to a unified scoring mechanism and a unified model that seamlessly bridges retrieval and reranking.
The contribution is therefore not PRF or RankNet in isolation, but the learned conversion of listwise context into a ranking-oriented query embedding while keeping document embeddings context-independent and reusable across retrieval and reranking.

We then introduce \model (\textbf{E}fficient \textbf{E}mbedding-based \textbf{Rank}ing or \textbf{Embedding-to-Rank}) with a two-stage training process, shown in Figure~\ref{fig:cover}: we first train an embedding model with contrastive learning, then jointly optimize a contrastive and a RankNet loss~\citep{burges2005ranknet} using the listwise prompt as a pseudo query. At the inference time, reranking reduces to cosine similarity between document embeddings and the query embedding augmented by the listwise prompt, which avoids time-consuming autogressive decoding while preserving the full query--document interaction.

We evaluate \model on popular reranking and embedding benchmarks. Experimental results demonstrate that our model achieves state-of-the-art reranking performance on BEIR~\citep{thakur2021beir} and exhibits a strong performance on reasoning-intensive benchmark BRIGHT~\citep{hongjinbright}, while notably improving inference efficiency. Additionally, trained solely on public data, our model preserves a competitive embedding performance on MTEB~\citep{muennighoff2022mteb}, demonstrating the effectiveness of unifying retrieval and reranking.\footnote{Code and models are available at \url{https://Alibaba-NLP.github.io/E2Rank}.}

\section{Related Work}

\paragraph{LLM-based Document Reranking}
LLMs such as GPT-4~\citep{openai2024gpt4} and Qwen-3~\citep{yang2025qwen3} have advanced document ranking across multiple benchmarks~\citep{sun2023rankgpt, zhu2023llm4ir, chen2024tourrank}, with prompting strategies typically falling into pointwise~\citep{liang2022holistic,sachan2022qg,zhang2023rankinggpt,liu2024demorank}, pairwise~\citep{qin2023pairwise}, and listwise paradigms~\citep{sun2023rankgpt,pradeep2023rankzephyr,liu2024sliding}. Among them, listwise methods are most effective by jointly considering the candidate set, and recent work further improves them through better prompting or output formats~\citep{reddy2024first,liu2025perank, chen2024attention, zhang2025query}.

\paragraph{Text Embedding Models}
Text embeddings map queries and documents into a shared semantic space and have become a foundation of modern search systems. Built on pre-trained encoders such as BERT~\citep{devlin2018bert} and T5~\citep{raffel2020t5}, models like GTE~\citep{li2023towards}, E5~\citep{wang2022e5}, and BGE~\citep{xiao2023bge} substantially improved retrieval via large-scale contrastive learning~\citep{karpukhin2020dense, ni2021gtr,zhao2024dense}. More recently, LLM-based embeddings~\citep{behnamghaderllm2vec, wang2023e5mistral, lee2025nv, zhang2025qwen3embedding} exploit stronger semantic understanding and instruction following~\citep{su2022instructor, li2024bgeicl}, and GritLM~\citep{muennighoff2024grit} further unifies embedding and generation. In contrast, our work unifies embedding and \emph{listwise} reranking within a single embedding model.

\paragraph{Pseudo Relevance Feedback for Dense Retrieval}
Pseudo Relevance Feedback (PRF) is a classical query-expansion technique that treats top-retrieved documents as relevant and uses them to refine the query~\citep{xu1996query, manning2008introduction}. PRF has been adapted to dense retrievers, e.g., ANCE-PRF~\citep{yu2021anceprf} learns a query encoder over the query and top documents, though it is less robust for stronger models~\citep{li2022improving, li2023pseudo}; other works apply PRF to pointwise cross-encoder rerankers via keyword expansion~\citep{li2024query, weller2024generative}. We instead interpret PRF within LLM-based \emph{listwise} reranking and show its effectiveness.

\section{Methodology}

We first review embedding-based retrieval and LLM-based listwise reranking, then present our key insight: listwise prompts can be treated as \emph{pseudo relevance feedback queries}, then the cosine similarity of embeddings could be a unified ranking function, leading to a unified model \model. Finally, we detail the training of \model.

\subsection{Preliminary}

For a LLM-based decoder-only text embedding model $f$ and any document $d$, we append the special end-of-sequence token [EOS] at the end of the input sequence, and the hidden state at the position of [EOS] from the final decoder layer is taken as the sequence embedding: $\bm{e}^{d} = f(d, \text{[EOS]})[-1]$. Further, given a query $q$, we append the instruction $I$ in front of the query to ensure its instruction-following abilities~\citep{su2022instructor} and obtain the embedding $\bm{e}^{q} = f(I, q, \text{[EOS]})[-1]$. The relevance between the query and the document is measured by the cosine similarity between their corresponding embeddings, denoted as $s(q, d) = \cos{(\bm{e}^{q}, \bm{e}^{d})}$.

While the embedding model encodes the semantic information of a \textit{single} document in the embedding space, it has not been optimized to capture nuanced differences between \textit{multiple} documents.
In contrast, LLM-based listwise rerankers (e.g., RankGPT~\citep{sun2023rankgpt}) use a \emph{listwise prompt} that includes the query and the entire candidate set, formulated as $\hat{q} = (I, d_1, ..., d_k, q)$, where $\{d_i\}_{i=1}^k$ is candidate documents set. The model is then asked to output a permutation of document identifiers in text form (e.g., ``[2] $>$ [1] $>$ [3]...'') of the documents in decreasing order of relevance. Although effective, this approach relies on autoregressive decoding of the document identifiers and incurs high latency, while recent work suggests that the listwise prompt itself---rather than autoregressive decoding---contribute the most to the ranking effectiveness~\citep{chen2024attention, zhang2025query}.

\subsection{Listwise Prompts as Pseudo Relevance Feedback Query}

Inspired by these observations, we propose to reinterpret the listwise prompt as a \emph{pseudo-relevance feedback (PRF)} query. Therefore, we can formulate the listwise reranking and retrieval in a \emph{unified} framework. Specifically, instead of generating a ranking list auto-regressively, we start from an embedding model and use the cosine similarity of embeddings as a unified ranking function for both retrieval and reranking. Formally, for the listwise prompt, we obtain its embedding
\begin{equation}
    \bm{e}^{\hat{q}} = f(I, d_1, ..., d_k, q)[-1],
\end{equation}
and compute $s(\hat{q}, d_i) = \cos{(\bm{e}^{\hat{q}}, \bm{e}^{d_i})}$ as the score for reranking.
The instructions we use is similar to ``\textit{Given a query and some relevant documents, rerank the documents}'', detailed in the Appendix~\ref{app:implementation}. It should be noted that, different from text embedding, we apply chat templates for listwise prompt.

This design allows us to exploit listwise information without sacrificing inference efficiency: the listwise prompt provides contextual PRF signals that let the model refine the query representation by implicitly leveraging document--document and query--document relationships, while both retrieval and reranking reduce to cosine similarity in a shared embedding space with reusable document embeddings.

\subsection{Training the Unified Embedding and Listwise Reranking Model}

We propose training \model in two stages: first, training an embedding model, then endowing it with listwise reranking capacity.

\paragraph{Stage I}
We start from training an LLM-based decoder-only text embedding model. In the training process, we employ standard contrastive learning to align relevant query--document pairs while pushing apart irrelevant ones. Specifically, for a training query $q_i$, there is one positive document $d_i^+$ and a set of negative documents $D^-$.
Given a batch of $N$ instances, we minimize the InfoNCE loss~\citep{izacard2021contriever}:
\begin{equation}
\resizebox{\columnwidth}{!}{$\displaystyle
    \mathcal{L}_{\text{InfoNCE}} = - \frac{1}{N} \sum_{i=1}^N \log \frac{e^{s(q_i, d_i^+)/\tau}}{e^{s(q_i, d_i^+)/\tau} + \sum_{d_j\in D^-} e^{s(q_i, d_j)/\tau}},$}
\end{equation}
We set the temperature $\tau$ to 0.03, following the training recipe used for the base embedding models, and do not tune it. This embedding training stage ensures that the base embedding model learns strong semantic representations suitable for large-scale retrieval.

\paragraph{Stage II} To incorporate the listwise reranking capabilities into the embedding model, we continue training the model using a multi-task learning framework. Basically, we include the contrastive learning with InfoNCE loss to maintain the embedding capacity of the model and a new learning-to-rank loss function, RankNet~\citep{burges2005ranknet} loss, which is a pairwise loss that measures the correctness of relative orders, for listwise ranking ability. The RankNet loss is defined as follows:
\begin{equation}
\begin{split}
    \mathcal{L}_{\text{RankNet}} = \frac{1}{N} \sum_{i=1}^N \sum_{d_j \in D} \sum_{d_k \in D} \mathbf{1}_{r_j < r_k} \\
    \cdot \log(1 + e^{s(q_i, d_j) / \tau - s(q_i, d_k) / \tau}),
\end{split}
\end{equation}
where $D$ is the same set of documents as used in contrastive learning (including both positive and negative). We fix $\tau=0.1$ for the RankNet loss across all model sizes and benchmarks; this scale places the small cosine-similarity differences in the sensitive range of the logistic loss. $r_j$ is the rank of document $d_j$ among $D$, and the smaller the rank, the more relevant. For example, $r_j = 2$ means $d_j$ ranks second among $|D|$ documents.
Following~\citep{sun2023rankgpt, pradeep2023rankzephyr}, we can leverage a powerful LLM to generate the full ranking permutation and obtain a set of pairwise relative relevance orders.
The final training objective of stage II combines retrieval and reranking losses:
\begin{equation}
    \mathcal{L} = \mathcal{L}_{\text{InfoNCE}} + \lambda \mathcal{L}_{\text{RankNet}},
\end{equation}
where $\lambda$ balances the two tasks. We set $\lambda=2.0$ based on a sensitivity sweep; performance varies by less than 1.0 NDCG@10 across $\lambda\in\{0.5,1.0,2.0,3.0\}$ (Appendix~\ref{app:implementation}).

\begin{table*}[t]
\providecommand{\sigbetter}[1]{\textcolor{green!45!black}{#1\textsuperscript{$\uparrow$}}}
\providecommand{\sigworse}[1]{\textcolor{red!70!black}{#1\textsuperscript{$\downarrow$}}}
\setlength{\tabcolsep}{3.8pt}
\centering
\scriptsize
\caption{Performance comparison on TREC DL and BEIR across LLMs. We \textbf{bold} the best result for each base LLM. The second column denotes training data. In the difference rows, \textcolor{green!45!black}{green \textsuperscript{$\uparrow$}}/\textcolor{red!70!black}{red \textsuperscript{$\downarrow$}} marks \model as significantly better/worse than the same-size RankQwen3 under a per-query paired two-sided $t$-test ($p<0.05$); unmarked differences are not significant.}
\label{tab:beir_comparison}
\begin{tabular}{l|l|cc|cccccccc|c}
\toprule
\textbf{Model}                               & \textbf{Data}                                & \textbf{DL19}  & \textbf{DL20}  & \textbf{COVID} & \textbf{NFCorpus} & \textbf{Touch\'{e}} & \textbf{DBPedia} & \textbf{SciFact} & \textbf{Signal} & \textbf{News}  & \textbf{Robust} & \textbf{Avg.}  \\
\midrule
BM25                                         & -                                            & 50.58          & 47.96          & 59.47          & 30.75             & 44.22           & 31.80            & 67.89            & 33.05           & 39.52          & 40.70           & 43.43          \\
\midrule
RankQwen3-0.6B                               & MS.                                           & 69.11          & 67.74          & 78.35          & 36.41             & 37.54           & 39.19            & 71.01            & 30.96           & 44.43          & 46.31           & 48.03          \\
\rowcolor{green!12}
\model-0.6B                                  & MS.                                           & 70.78          & \textbf{70.55} & \textbf{80.03} & 37.63             & 36.60           & 41.90            & 73.19            & \textbf{35.66}  & 51.17          & 49.70           & 50.74          \\
                                             &                                               & \sigbetter{+1.67} & \sigbetter{+2.81} & +1.68 & \sigbetter{+1.22} & \sigworse{-0.94} & \sigbetter{+2.71} & \sigbetter{+2.18} & \sigbetter{+4.70} & \sigbetter{+6.74} & \sigbetter{+3.39} & +2.71 \\
\rowcolor{varysorange!25}
\model-0.6B                                  & Mixed                                         & \textbf{70.84} & 70.15          & 79.17          & \textbf{38.60}    & \textbf{41.91}  & \textbf{41.96}   & \textbf{73.43}   & 35.26           & \textbf{52.75} & \textbf{53.67}  & \textbf{52.09} \\
                                             &                                               & \sigbetter{+1.73} & \sigbetter{+2.41} & +0.82 & \sigbetter{+2.19} & \sigbetter{+4.37} & \sigbetter{+2.77} & \sigbetter{+2.42} & \sigbetter{+4.30} & \sigbetter{+8.32} & \sigbetter{+7.36} & +4.07 \\
\midrule
RankQwen3-4B                                 & MS.                                           & \textbf{72.36} & 69.83          & 83.91          & \textbf{39.88}    & 32.66           & \textbf{43.91}   & 76.37            & 32.15           & 50.81          & 59.36           & 52.38          \\
\rowcolor{green!12}
\model-4B                                    & MS.                                           & 70.67          & \textbf{71.05} & \textbf{84.90} & 39.32             & 35.44           & 43.66            & \textbf{77.69}   & 34.21           & 51.22          & 57.49           & 52.99          \\
                                             &                                               & \sigworse{-1.69} & +1.22 & +0.99 & -0.56 & \sigbetter{+2.78} & -0.25 & +1.32 & \sigbetter{+2.06} & +0.41 & \sigworse{-1.87} & +0.61 \\
\rowcolor{varysorange!25}
\model-4B                                    & Mixed                                         & 70.44          & 70.64          & 83.30          & 39.20             & \textbf{43.16}  & 42.95            & 77.19            & \textbf{34.48}  & \textbf{52.71} & \textbf{60.16}  & \textbf{54.14} \\
                                             &                                               & \sigworse{-1.92} & +0.81 & -0.61 & -0.68 & \sigbetter{+10.50} & -0.96 & +0.82 & \sigbetter{+2.33} & \sigbetter{+1.90} & +0.80 & +1.76 \\
\midrule
RankQwen3-8B                                 & MS.                                           & \textbf{73.15} & 70.75          & 85.37          & \textbf{40.05}    & 31.73           & \textbf{45.44}   & \textbf{78.96}   & 32.48           & 52.36          & \textbf{60.72}  & 53.39          \\
\rowcolor{green!12}
\model-8B                                    & MS.                                           & 71.66          & 70.87          & \textbf{85.43} & 39.57             & 36.59           & 44.26            & 78.17            & 33.52           & \textbf{55.36} & 57.95           & 53.86          \\
                                             &                                               & -1.49 & +0.12 & +0.06 & -0.48 & \sigbetter{+4.86} & \sigworse{-1.18} & -0.79 & +1.04 & \sigbetter{+3.00} & \sigworse{-2.77} & +0.47 \\
\rowcolor{varysorange!25}
\model-8B                                    & Mixed                                         & 72.95          & \textbf{71.16} & 84.09          & 39.08             & \textbf{42.06}  & 43.44            & 77.49            & \textbf{34.01}  & 54.25          & 60.34           & \textbf{54.35} \\
                                             &                                               & -0.20 & +0.41 & -1.28 & -0.97 & \sigbetter{+10.33} & \sigworse{-2.00} & \sigworse{-1.47} & \sigbetter{+1.53} & \sigbetter{+1.89} & -0.38 & +0.96 \\ 
\bottomrule
\end{tabular}
\end{table*}

\begin{table*}[t]
\centering
\caption{Performance comparison on BRIGHT benchmarks across LLMs. We \textbf{bold} the
best performance for each task and {\ul underline} the second best.}
\label{tab:bright}
\scriptsize
\setlength{\tabcolsep}{4.5pt}
\begin{tabular}{@{ }l|ccccccc|cc|ccc|c@{ }}
\toprule
\multirow{2}{*}{\textbf{Model}} &
  \multicolumn{7}{c|}{\textbf{StackExchange}} &
  \multicolumn{2}{c|}{\textbf{Coding}} &
  \multicolumn{3}{c|}{\textbf{Theorem-based}} &
  \multicolumn{1}{l}{\multirow{2}{*}{\textbf{Avg.}}} \\ \cmidrule(lr){2-13}
 &
  \textbf{Bio.} &
  \textbf{Econ.} &
  \textbf{Earth.} &
  \textbf{Psy.} &
  \textbf{Rob.} &
  \textbf{Stack.} &
  \textbf{Sus.} &
  \textbf{Pony.} &
  \textbf{LC.} &
  \textbf{AoPS} &
  \textbf{TheoQ.} &
  \textbf{ThoT.} &
  \multicolumn{1}{l}{} \\
\midrule
ReasonIR         & 43.5          & 32.8           & 43.0            & 38.9          & 21.1          & 30.6            & 27.3          & 31.6           & 19.6          & 7.3           & 36.7            & 34.1           & 30.5          \\ 
\midrule
RankT5 (3B)      & 11.4          & 22.1           & 10.9            & 13.6          & 11.4          & 11.4            & 16.0          & 27.5           & 38.1          & 9.2           & 18.3            & 9.5            & 16.6          \\
RankZephyr       & 19.9          & 17.4           & 12.4            & 34.9          & 24.7          & 13.4            & 22.3          & 29.3           & 32.4          & 6.1           & 29.0            & 30.1           & 22.6          \\ 
\midrule
Rank-R1 (7B)     & 39.3          & 28.1           & 23.9            & 30.0          & 17.3          & 18.1            & 33.2          & 18.6           & 15.0          & 4.2           & 25.4            & 35.7           & 24.1          \\
Rank-R1 (14B)    & 27.4          & 38.7           & 23.1            & 44.5          & 37.1          & 27.8            & 36.8          & 21.3           & 19.2          & 8.8           & 31.7            & 39.5           & 29.7          \\
Rank1 (7B)       & 44.1          & 33.5           & 21.8            & 30.0          & 15.0          & 22.1            & 28.5          & 11.8           & 21.7          & 1.2           & 26.2            & 36.2           & 24.3          \\
Rearank (7B)     & 35.3          & 29.8           & 25.5            & 35.7          & 19.1          & 20.1            & 32.9          & 29.9           & 20.2          & 6.2           & 36.7            & 38.3           & 27.5          \\
JudgeRank (8B)   & 37.1          & 27.2           & 19.2            & 28.6          & 11.6          & 19.9            & 22.5          & 10.2           & 10.2          & 3.6           & 22.9            & 29.4           & 20.2          \\
ERank (4B)       & 42.1          & 42.5           & 26.3            & 36.4          & 20.8          & 27.3            & 33.2          & {\ul 31.7}     & 21.8          & {\ul 10.9}    & 32.8            & 40.6           & 30.5          \\
ERank (14B)      & 46.6          & 42.5           & 25.2            & 37.3          & 19.6          & 30.2            & 34.6          & \textbf{31.9}  & 25.6          & 10.5          & 32.4            & \textbf{45.0}  & 31.8          \\
ReasonRank (7B)  & 35.1          & {\ul 47.8}     & 31.2            & \textbf{56.7} & \textbf{47.8} & {\ul 32.5}      & \textbf{40.9} & 23.2           & 25.0          & 7.7           & {\ul 39.5}      & {\ul 41.8}     & \textbf{35.7} \\ 
\midrule
RankQwen3-0.6B   & 44.7          & 38.7           & 28.4            & 40.4          & 20.5          & 26.1            & 28.5          & 19.9           & 29.1          & 6.8           & 35.8            & 30.5           & 29.1          \\
\rowcolor{green!12}
\model-0.6B (MS.)& 41.2          & 46.5           & 30.9            & 34.3          & 24.2          & 24.0            & 27.4          & 9.4            & 35.6          & 10.5          & 30.4            & 27.2           & 28.5                                               \\
\rowcolor{varysorange!25}
\model-0.6B      & 44.1          & 46.5           & 31.0            & 40.8          & 26.1          & 30.6            & 30.6          & 11.7           & \textbf{38.5} & 8.0           & 35.9            & 28.0           & 31.0          \\ 
\midrule
RankQwen3-4B     & 47.0          & 44.2           & 25.2            & 44.7          & 24.1          & 29.7            & 41.1          & 22.6           & 22.0          & 9.0           & 38.2            & 36.0           & 32.0          \\
\rowcolor{green!12}
\model-4B (MS.)  & 42.9          & 44.4           & 30.5            & 39.1          & 27.4          & 25.4            & 32.9          & 7.7            & 38.3          & \textbf{11.4} & 38.4            & 35.5           & 31.2                                               \\
\rowcolor{varysorange!25}
\model-4B        & 47.6          & 46.7           & {\ul 31.8}      & 43.1          & 26.8          & 31.4            & 34.6          & 8.6            & {\ul 38.4}    & 8.2           & \textbf{39.8}   & 31.6           & 32.4          \\ 
\midrule
RankQwen3-8B     & \textbf{49.5} & 44.2           & 30.4            & {\ul 44.9}    & 24.9          & 26.1            & {\ul 39.6}    & 18.8           & 20.8          & 7.6           & 39.0            & 37.9           & 32.0          \\
\rowcolor{green!12}
\model-8B (MS.)  & 45.0          & \textbf{48.1}  & 31.9            & 38.5          & {\ul 29.3}    & 32.2            & 36.4          & 10.3           & 36.0          & 10.6          & 37.9            & 36.6           & 32.7                                               \\
\rowcolor{varysorange!25}
\model-8B        & {\ul 49.2}    & 47.2           & \textbf{32.3}   & 44.7          & 28.2          & \textbf{32.9}   & 38.4          & 10.6           & 36.2          & 8.2           & 38.2            & 33.4           & {\ul 33.4}    \\ 
\bottomrule
\end{tabular}
\end{table*}

\section{Experiments}

\subsection{Experimental Setup}\label{sec:exp_setup}

\paragraph{Base LLMs} We conduct our experiments with open-weight, instruction-tuned LLMs from the Qwen3 family~\citep{yang2025qwen3} across different sizes, including 0.6B, 4B, and 8B.

\paragraph{Training Datasets} At Stage I, we use the public portion of the E5 training dataset~\citep{wang2023e5mistral} with roughly 1.5 million samples, curated by~\cite{springer2025repetition}. For the second stage training, we compare two different training datasets. First, we use the MS MARCO training data provided by~\cite{pradeep2023rankzephyr} with 40K samples. The second training dataset we use is a mixture, consisting of some of the retrieval datasets from the Stage I, as well as 2 additional public Chinese retrieval datasets from BGE-M3 training dataset~\citep{chen2024bgem3}. We further sample instances from these datasets and construct hard negatives for each query, resulting in about 87k training samples each with 1 query, 1 positive, and 15 negatives. We leverage Qwen3-32B to label the ranking permutations and treat them as noisy supervision rather than ground truth: gold-positive-first agreement is only slightly above 50\% on MS MARCO, although the zero-shot teacher reaches 53.37 average NDCG@10 on the BEIR subset and 23.8 on BRIGHT. Appendix~\ref{app:dataset} provides the full analysis; we did not conduct a separate human-label study.

\paragraph{Implementation Details} We train the embedding model with full parameters for 1 epoch with a batch size of 512, using a learning rate of 5e-6. At the second stage, we continue to train the model for about 700 steps with a batch size of 128, and the number of negatives is 15. We provide other hyperparameters in Appendix~\ref{app:implementation}.

\subsection{Reranking Performance}\label{sec:reranking}

\paragraph{Datasets} Following the established listwise-reranking protocol of RankGPT~\citep{sun2023rankgpt}, we use TREC DL~\citep{craswell2020trecdl} and eight BEIR datasets~\citep{thakur2021beir}: TREC-COVID, NFCorpus, Touch\'{e}2020, DBPedia, SciFact, Signal1M, TREC News, and Robust04. This subset spans diverse domains while keeping top-100 sliding-window evaluation of generative baselines tractable and directly comparable with prior listwise work. We also evaluate \model on the reasoning-intensive BRIGHT benchmark~\citep{hongjinbright}.
We use BM25 as the first-stage retriever for TREC DL and BEIR and use ReasonIR~\citep{shao2025reasonir} with GPT4 reason-query for BRIGHT. For all benchmarks, we rerank the top-100 candidate documents and use NDCG@10 as the metric.

\paragraph{Baselines} For a controlled comparison, we fine-tune the same Qwen3 backbones on the RankZephyr data~\citep{pradeep2023rankzephyr}, yielding the generative listwise baseline RankQwen3 (Appendix~\ref{app:rankqwen}). RankQwen3 uses sliding windows of size 20 and step 10, whereas \model feeds only the top-20 documents to its listwise prompt and uses the resulting embedding to rerank the top-100.

For broader comparisons on TREC DL and BEIR, we also report cross-encoders monoBERT~\citep{nogueira2019monobert}, monoT5~\citep{nogueira2020monot5}, and RankT5~\citep{zhuang2023rankt5}, as well as listwise LLM-based rerankers ListT5~\citep{yoon2024listt5}, RankZephyr~\citep{pradeep2023rankzephyr}, FirstZephyr~\citep{reddy2024first}, and RankGPT~\citep{sun2023rankgpt}. On BRIGHT, we compare with rerankers below 14B parameters, including Rank-R1~\citep{zhuang2025rank}, Rank1~\citep{weller2025rank1}, JudgeRank~\citep{niu2024judgerank}, Rearank~\citep{zhang2025rearank}, ERank~\citep{cai2025erank}, and ReasonRank~\citep{liu2025reasonrank}. Only RankGPT and JudgeRank are zero-shot; the remaining methods are fine-tuned, and most reasoning rerankers use RL.

\paragraph{\model improves average effectiveness over RankQwen3.} As shown in Table~\ref{tab:beir_comparison}, \model improves the average over the same-backbone RankQwen3 at every scale, including $+2.71$ NDCG@10 at 0.6B under matched training data. The gains concentrate on the out-of-domain BEIR datasets; on TREC-DL, results are mixed and several differences at 4B/8B are not statistically significant. Table~\ref{tab:beir_comparison} reports per-query paired two-sided t-tests ($p<0.05$) for these controlled comparisons. The significance pattern is strongest at 0.6B: with mixed training data, \model is significantly better on nine of ten datasets and is never significantly worse. At 4B/8B, significant gains are more dataset-specific---concentrating on Touch\'{e}2020 and often Signal1M or TREC News---while most TREC-DL differences are not significant and a few BEIR datasets regress significantly.

\paragraph{\model achieves competitive rerank accuracy across other strong baselines.} Table~\ref{tab:beir_all} presents broader comparisons on TREC DL and BEIR. \model outperforms fine-tuned pointwise rerankers such as monoBERT and monoT5, and surpasses strong listwise baselines like RankZephyr and ListT5 on BEIR. Our fine-tuned 8B model attains the top DL20 score (71.16) and the highest BEIR average (54.35), outperforming even the much larger zero-shot RankGPT-4o.

\begin{table}[t]
\centering
\scriptsize
\caption{Performance comparison across broader baselines. The best result of each benchmark is \textbf{bolded}, and the second best is {\ul underlined}. RankGPT results are from~\citep{liu2024sliding}, which reranks BM25 top-100 under the same protocol rather than using the cascaded setup of the original RankGPT paper.}
\label{tab:beir_all}
\begin{tabular}{@{}lccc@{}}
\toprule
\textbf{Model}  & \textbf{DL19}  & \textbf{DL20}  & \textbf{BEIR Avg.} \\
\midrule
BM25 & 50.58 & 47.96 & 43.43 \\
\midrule
\multicolumn{4}{l}{\textit{Fine-tuned Pointwise Reranker}}             \\
\midrule
monoBERT (340M) & 70.50          & 67.28          & 47.16              \\
monoT5 (3B)     & 71.83          & 68.89          & 51.36              \\
RankT5 (3B)     & 72.50          & 70.40          & 52.50              \\
\midrule
\multicolumn{4}{l}{\textit{Fine-tuned Listwise Reranker}}              \\
\midrule
ListT5 (3B)     & 71.80          & 69.10          & 53.00              \\
FirstZephyr     & 69.50          & 65.94          & 48.68              \\
RankZephyr      & 73.39          & 70.02          & 51.15              \\
\midrule
\multicolumn{4}{l}{\textit{Zero-shot Listwise Reranker}}               \\
\midrule
RankQwen3 (14B) & {\ul 74.19}    & 69.10          & 53.67              \\
RankGPT-4o      & \textbf{74.78} & 69.52          & 53.09              \\
RankGPT-4o-mini & 72.36          & 67.30          & 51.16              \\
\midrule
\multicolumn{4}{l}{\textit{Ours}}                                               \\
\midrule
\model-0.6B     & 70.84          & 70.15          & 52.09              \\
\model-4B       & 70.44          & {\ul 70.64}    & {\ul 54.14}        \\
\model-8B       & 72.95          & \textbf{71.16} & \textbf{54.35}     \\
\bottomrule
\end{tabular}
\end{table}

\begin{table*}[t]
\centering
\scriptsize
\caption{Performance comparison on MTEB. Note that some baselines are trained with non-public data, and we only report the version trained on public data, marked using *. The best results for each subtask are highlighted in \textbf{bold}, and the second-best results are {\ul underlined}.}
\label{tab:mteb_v1_comparison}
\begin{tabular}{@{}l|ccccccc|c@{}}
\toprule
\textbf{Categorias $\rightarrow$}                & \textbf{Retr.} & \textbf{Rerank.} & \textbf{Clust.} & \textbf{PairClass.} & \textbf{Class.} & \textbf{STS}  & \textbf{Summ.} & \textbf{Avg.} \\
\# of datasets $\rightarrow$                     & 15             & 4                & 11              & 3                   & 12              & 10            & 1              & 56            \\ \midrule
Instructor-xl~\citep{su2022instructor}           & 49.26          & 57.29            & 44.74           & 86.62               & 73.12           & 83.06         & \textbf{32.32} & 61.79         \\
BGE$_{\text{large-en-v1.5}}$~\citep{xiao2023bge} & 54.29          & 60.03            & 46.08           & 87.12               & 75.97           & 83.11         & 31.61          & 64.23         \\
GritLM$_{\text{Mistral-7b-v1}}$*~\citep{muennighoff2024grit} & 53.10 & \textbf{61.30}  & \textbf{48.90}  & 86.90               & 77.00           & 82.80         & 29.40          & 64.70         \\
E5$_{\text{Mistral-7b-v1}}$*~\citep{wang2023e5mistral} & 52.78    & {\ul 60.38}      & {\ul 47.78}     & \textbf{88.47}      & 76.80           & 83.77         & {\ul 31.90}    & 64.56         \\
Echo$_{\text{Mistral-7b-v1}}$~\citep{springer2025repetition} & 55.52 & 58.14          & 46.32           & 87.34               & \textbf{77.43}  & 82.56         & 30.73          & 64.68         \\
LLM2Vec$_{\text{Mistral-7B}}$~\citep{behnamghaderllm2vec} & 55.99 & 58.42             & 45.54           & {\ul 87.99}         & 76.63           & {\ul 84.09}   & 29.96          & 64.80         \\
LLM2Vec$_{\text{Meta-LLaMA-3-8B}}$~\citep{behnamghaderllm2vec} & 56.63 & 59.68         & 46.45           & 87.80               & 75.92           & 83.58         & 30.94          & {\ul 65.01}   \\
BGE-en-icl$_{\text{Mistral-7b-v1}}$*~\citep{li2024bgeicl} (zero-shot) & \textbf{59.59} & 56.85 & 42.61      & 87.87               & 75.47           & 83.30         & 29.52          & 64.67         \\
\midrule 
\model-0.6B (w/ only Stage I)                    & 48.07          & 56.16            & 42.38           & 82.47               & 72.05           & 80.90         & 29.84          & 60.05         \\ 
\model-0.6B                                      & 51.74          & 55.97            & 40.85           & 83.93               & 73.66           & 81.41         & 30.90          & 61.25         \\ \midrule
\model-4B (w/ only Stage I)                      & 54.36          & 59.30            & 44.62           & 84.36               & 76.11           & 82.31         & 29.33          & 63.61         \\ 
\model-4B                                        & 55.33          & 59.10            & 44.27           & 87.14               & {\ul 77.08}     & 84.03         & 30.06          & 64.47         \\ \midrule
\model-8B (w/ only Stage I)                      & 55.31          & 55.73            & 45.84           & 85.23               & 75.69           & 83.23         & 29.66          & 64.26         \\ 
\model-8B                                        & {\ul 56.89}    & 59.58            & 44.75           & 86.96               & 76.81           & \textbf{84.52}& 30.23          & \textbf{65.03}\\ 
\bottomrule
\end{tabular}
\end{table*}

\paragraph{Efficiency Analysis.}
We measure latency on DL20, TREC-COVID, DBPedia, and Robust04 on a single NVIDIA A100 80G GPU using vLLM~\citep{kwon2023vllm}. As shown in Table~\ref{tab:latency}, \model substantially reduces latency across all sizes, achieving up to $\sim 5\times$ speedup over RankQwen3 at 8B on TREC-COVID. It also requires no decoding and is faster than the single-token-decoding FirstZephyr-7B on every dataset, while supporting batch inference and reusable document embeddings.

\begin{table}[t]
\centering
\caption{Reranking latency per query (seconds) on a single NVIDIA A100 80G GPU. All methods rerank the top-100 candidates; \model latency includes document encoding.}
\label{tab:latency}
\scriptsize
\setlength{\tabcolsep}{2.2pt}
\begin{tabular}{@{}lcccc@{}}
\toprule
\textbf{Method} & \textbf{DL20} & \textbf{COVID} & \textbf{DBPedia} & \textbf{R04} \\
\midrule
Qwen3-0.6B (pointwise) & 0.28 & 0.40 & 0.28 & 0.38 \\
Qwen3-4B (pointwise)   & 1.01 & 1.39 & 0.99 & 1.20 \\
Qwen3-8B (pointwise)   & 1.77 & 2.32 & 1.74 & 2.31 \\
\midrule
RankQwen3-0.6B & 1.97 & 4.58 & 1.93 & 4.33 \\
RankQwen3-4B   & 4.23 & 11.25 & 4.16 & 11.03 \\
RankQwen3-8B   & 6.44 & 16.93 & 6.40 & 15.36 \\
FirstZephyr-7B & 3.81 & 5.26 & 3.66 & 5.18 \\
\midrule
\model-0.6B & 0.45 & 0.63 & 0.47 & 0.59 \\
\model-4B   & 1.65 & 2.17 & 1.60 & 2.12 \\
\model-8B   & 2.55 & 3.40 & 2.51 & 3.33 \\
\bottomrule
\end{tabular}
\end{table}

\paragraph{\model demonstrates strong performance on the BRIGHT benchmark.} On the challenging BRIGHT benchmark, \model delivers robust performance, as shown in Table~\ref{tab:bright}. Without any RL or reasoning process, \model-8B attains a highly competitive average score of 33.4, surpassing RankQwen3 and most reasoning rerankers and only underperforming ReasonRank trained on synthetic reasoning data, validating the strong generalization capabilities.

\begin{table}[t]
    \centering
    \caption{End-to-end retrieval and reranking performance (NDCG@10). Qwen3E denotes Qwen3-Embedding; external first-stage retrievers are reranked with \model-0.6B.}
    \label{tab:end2end}
    \scriptsize
    \setlength{\tabcolsep}{4pt}
    \begin{tabular}{@{}l|ccc@{}}
    \toprule
    \textbf{Pipeline} & \textbf{DL20}  & \textbf{BEIR} & \textbf{BRIGHT}  \\ \midrule
 BM25 retrieval                         & 47.96 & 43.43 & 13.70 \\
 \quad + \model-0.6B reranking          & 70.15 & 52.09 & 20.20 \\ \midrule
 Qwen3E-0.6B retrieval                  & 66.69 & 48.35 & 14.40 \\
 \quad + \model-0.6B reranking          & 73.77 & 52.72 & 21.30 \\ \midrule
 \model-0.6B retrieval                  & 66.77 & 47.60 & 18.37 \\
 \quad + self-reranking                 & 74.40 & 50.66 & 22.58 \\ \midrule
 \model-4B retrieval                    & 74.00 & 52.11 & 27.84 \\
 \quad + self-reranking                 & 76.88 & 54.12 & \textbf{32.15} \\ \midrule
 \model-8B retrieval                    & 75.83 & 53.39 & 25.09 \\
 \quad + self-reranking                 & \textbf{78.02} & \textbf{55.08} & 31.00 \\
    \bottomrule
    \end{tabular}
\end{table}

\subsection{Embedding Ability}

\paragraph{Benchmark and Baselines} We evaluate \model on MTEB~\citep{muennighoff2022mteb} (English v1 with 56 datasets across 7 task types; v2 with 41 tasks for ablations), comparing with open-source embedding models trained on public data. This choice enables a fair comparison.

\paragraph{Results} Table~\ref{tab:mteb_v1_comparison} presents performance on MTEB (English, v1). When trained only on public data, \model retains strong embedding capabilities, and \model-8B is competitive with recent embedding models on average. Compared with contrastive learning alone, ranking supervision consistently improves retrieval tasks ($\uparrow 1.58$ for \model-8B), demonstrating the effectiveness of the ranking objective. The gains extend beyond retrieval to pair classification, classification, and STS across all model scales, whereas clustering shows a small trade-off. Here we use query-only embeddings rather than listwise prompts to measure pure embedding ability.

\subsection{Unified and End-to-End Retrieval and Reranking}

Table~\ref{tab:end2end} compares retrieval alone with end-to-end reranking for external and unified first-stage retrievers. \model improves every first stage, while Stage-II training largely preserves the 0.6B model's retrieval ability relative to the out-of-the-box Qwen3-Embedding model. The unified pipeline is strongest on DL20 and BRIGHT, whereas the Qwen3-Embedding cascade retains a small advantage over the size-matched unified pipeline on BEIR. The gains across sparse and dense first stages, particularly on BRIGHT for the larger unified models, indicate that listwise reranking is robust to the candidate generator and especially useful for reasoning-intensive queries. Scaling \model strengthens end-to-end performance.

\subsection{Ablation Study}

We evaluate the effectiveness of different training strategies and conduct ablation studies using the Qwen3-0.6B model on TREC DL20, BEIR, BRIGHT, and MTEB(eng, v2). The reranking settings and metrics are the same as in Section~\ref{sec:reranking}. The results shown in Table~\ref{tab:abl_training} indicate that the full training strategy achieves the best or highly competitive performance across all datasets, demonstrating the effectiveness of the integrated design. {For the last three lines, we use query-only embedding instead of listwise prompt for evaluation since they are not trained on it.}

\textbf{The first-stage contrastive learning is crucial for foundational query-document alignment and embedding ability.}
Its removal causes consistent performance degradation, especially on MTEB. This confirms that initial large-scale contrastive learning provides an essential foundation for subsequent ranking tasks.

\textbf{The RankNet loss is the most critical element for effective ranking.}
Removing the RankNet loss causes the most severe performance collapse, particularly on BEIR and BRIGHT. This underscores that the pairwise ranking objective is indispensable for learning complex relevance ordering patterns.

\textbf{The listwise prompts incorporating PRF documents substantially enhance ranking effectiveness.}
If retaining the RankNet loss but removing the listwise prompt, the ranking performance will still be greatly affected (last line). This indicates that the reranking ability is mainly from the listwise prompt with PRF signals, but not the richer training labels.

\begin{table}[t]
    \centering
    \caption{Ablation on different training strategies.}
    \label{tab:abl_training}
    \scriptsize
    \setlength{\tabcolsep}{3pt}
    \begin{tabular}{@{}l|cccc@{}}
    \toprule
                 & \textbf{DL20}  & \textbf{BEIR} & \textbf{BRIGHT} & \textbf{MTEB(v2)} \\ \midrule
    \model-0.6B  & \textbf{70.15} & {\ul 52.09}        & \textbf{30.96} & {\ul 63.41}            \\ \midrule
    w/o Stage I  & {\ul 69.32}    & 51.33              & {\ul 30.66}    & 60.61                  \\
    w/o InfoNCE in Stage II   & 69.11          & \textbf{52.17}     & 29.99          & 61.92                  \\ \midrule
    w/ only Stage I & 63.55          & 46.31              & 15.30          & 62.40                  \\
    w/o RankNet in Stage II  & 66.50          & 49.24              & 22.40          & 63.31                  \\ \midrule
    w/o Listwise in Stage II & 66.29          & 49.93              & 22.69          & \textbf{63.66}         \\
    \bottomrule
    \end{tabular}
\end{table}

\subsection{Analysis}\label{sec:analysis}

In order to understand the reranking behaviors of \model, we conduct a further analysis.

\paragraph{Comparison with other PRF methods.}
We compare \model against two classical PRF-style baselines built on the model without listwise training (``w/o Listwise in Stage II'' in Table~\ref{tab:abl_training}): (i) a text-based listwise prompt, and (ii) a vector-based Rocchio-style fusion of query and document embeddings. As shown in Table~\ref{tab:prf}, neither baseline approaches \model and both can even hurt ranking quality, demonstrating that supervision is necessary for the model to learn how to leverage PRF signals. We provide further discussion in Appendix~\ref{app:prf}.

\begin{table}[t]
\centering
\scriptsize
\caption{Performance comparison across other PRF baselines. The best result of each benchmark is \textbf{bolded}.}
\label{tab:prf}
\begin{tabular}{l|ccc}
\toprule
                          & \textbf{DL20}  & \textbf{BEIR}  & \textbf{BRIGHT} \\
\midrule
\model-0.6B               & \textbf{70.15} & \textbf{52.09} & \textbf{30.96}  \\
\quad w/o Listwise Prompt & 65.25 & 49.46 & 21.50  \\
\midrule
w/o Listwise in Stage II  & 66.29 & 49.93 & 22.69  \\
 \quad + text-based PRF   & 56.57 & 46.52 & 29.62  \\
 \quad + vector-based PRF & 63.96 & 49.20 & 21.85  \\
\bottomrule
\end{tabular}
\end{table}

\begin{table*}[t]
\centering
\scriptsize
\caption{Reranking results using different first-stage retrievers.}
\label{tab:first_stage}
\begin{tabular}{l|cc|cc|cc|cc}
\toprule
 &
  \multicolumn{2}{c|}{\textbf{BGE-base}} &
  \multicolumn{2}{c|}{\textbf{Contriver}} &
  \multicolumn{2}{c|}{\textbf{SPLADE++ED}} &
  \multicolumn{2}{c}{\textbf{Qwen3E-0.6B}} \\ 
  &
  \textbf{DL19} &
  \textbf{DL20} &
  \textbf{DL19} &
  \textbf{DL20} &
  \textbf{DL19} &
  \textbf{DL20} &
  \textbf{DL19} &
  \textbf{DL20} \\ \midrule
First-stage Retrieval & 70.22 & 66.21 & 62.02 & 63.42 & 73.08 & 71.97 & 68.05 & 66.69 \\ \midrule
RankQwen3-0.6B        & 72.60 & 72.51 & 68.63 & 71.78 & 75.82 & 74.34 & 74.00 & 72.65 \\
\model-0.6B           & 74.53 & 73.97 & 71.81 & 74.52 & 76.04 & 77.82 & 74.83 & 73.42 \\ \midrule
RankQwen3-4B          & 72.71 & 76.31 & 70.89 & 76.06 & 75.56 & 74.78 & 72.42 & 73.29 \\
\model-4B             & 75.46 & 74.90 & 72.71 & 76.01 & 75.74 & 79.25 & 74.92 & 74.88 \\ \midrule
RankQwen3-8B          & 73.73 & 75.68 & 72.62 & 75.94 & 74.61 & 75.81 & 73.96 & 75.26 \\
\model-8B             & 74.15 & 76.40 & 73.77 & 75.04 & 77.37 & 80.08 & 74.97 & 75.24 \\ \bottomrule
\end{tabular}
\end{table*}

\begin{figure}[t]
\centering
\begin{minipage}[t]{0.48\columnwidth}
\centering
\includegraphics[width=0.95\textwidth]{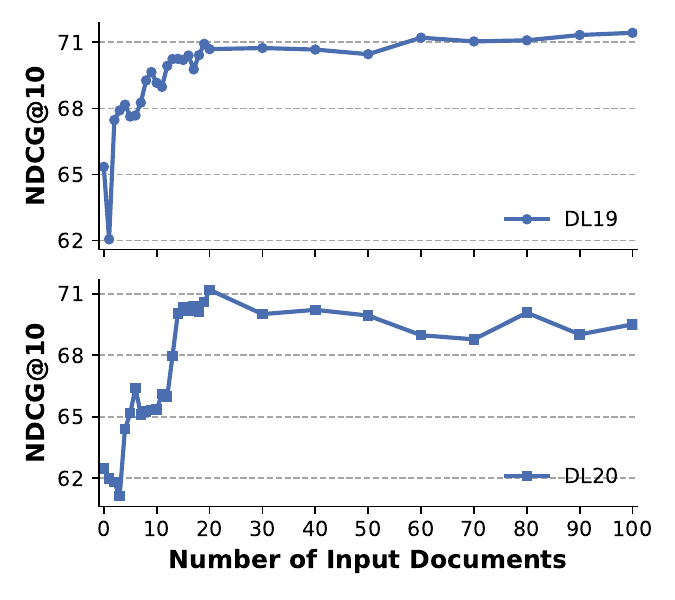}
\end{minipage}
\hfill
\begin{minipage}[t]{0.48\columnwidth}
\centering
\includegraphics[width=0.95\textwidth]{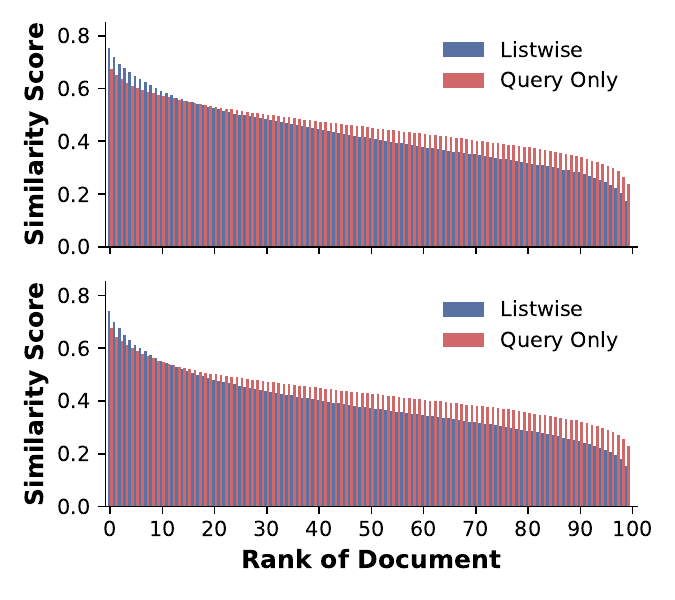}
\end{minipage}
\caption{Left: Trend of NDCG@10 changes with the number of input documents in listwise prompt. Right: Score distribution of using listwise (with 20 documents) and non-listwise prompts.}
\label{fig:num_docs_score}
\end{figure}

\paragraph{Influence of number of input documents in the listwise prompts.}
Figure~\ref{fig:num_docs_score} shows that when the number of input documents is small (less than 20), incorporating more documents into the listwise prompt consistently improves ranking performance. This trend can be interpreted as additional documents enriching the query with pseudo-relevance signals, allowing the model to capture the fine-grained relevance. Notably, the gains plateau after around 20 documents, indicating that the marginal benefit of adding more feedback signals diminishes once the prompt already captures sufficient relevance context, and may even bring negative benefits on different datasets.

\begin{figure}
    \centering
    \includegraphics[width=\columnwidth]{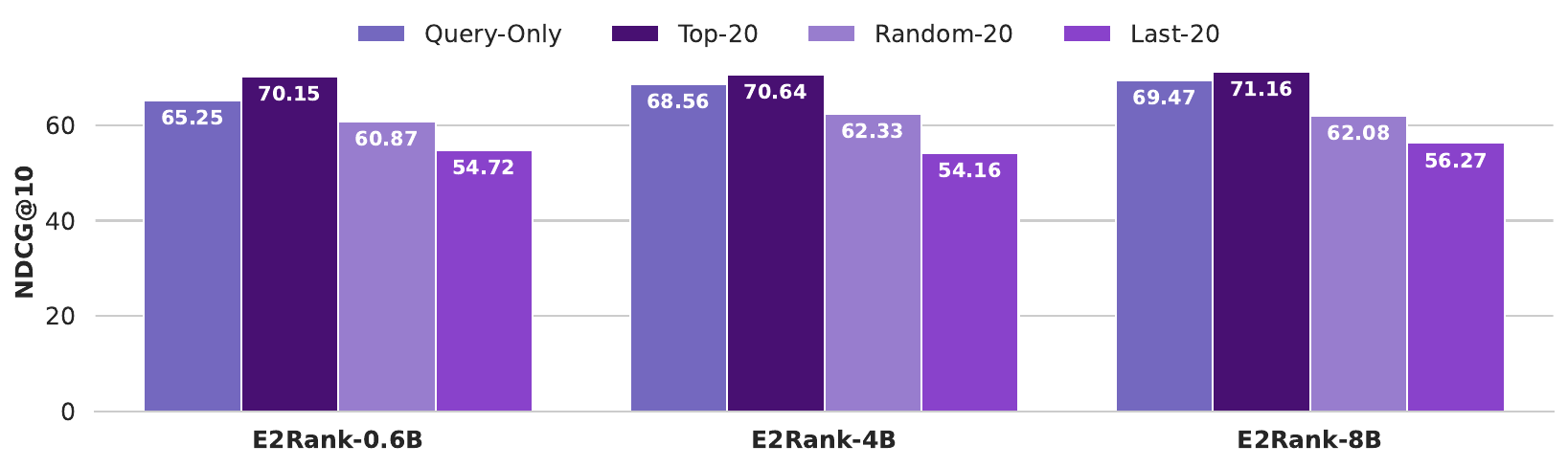}
    \caption{Results of selecting different documents as PRF on DL20.}
    \label{fig:diff_docs}
\end{figure}

\paragraph{Influence of the selection of documents.}
To evaluate the impact of document selection within the listwise prompt, we conducted an ablation study varying the composition of documents in four settings, as presented in Figure~\ref{fig:diff_docs}. Across all model sizes, using the Top-20 retrieved documents consistently yielded the best ranking performance. In contrast, Random-20 and Last-20 settings significantly reduced performance, even falling below the query-only baseline. This shows that the benefit of document context comes not from adding more text, but from leveraging highly relevant documents. Conversely, low-quality or off-topic documents introduce noise and conflicting signals, leading to deterioration in ranking quality.

\paragraph{Similarity score distribution.}
Figure~\ref{fig:num_docs_score} further analyzes how this pseudo relevance feedback affects the ranking behavior by comparing similarity score distributions between listwise and non-listwise settings. Specifically, we sort the reranking scores of 100 documents from high to low, and take the average of all queries for the rank position. We can see that the listwise prompts yield consistently higher similarity scores for top-ranked documents while maintaining a steeper decline for lower-ranked ones, suggesting sharper discrimination between relevant and irrelevant documents. In contrast, the query-only setting produces a flatter score distribution. This demonstrates that listwise prompts with PRF enhance \model's ability to allocate higher scores to truly relevant documents.

\paragraph{Influence of different first-stage retrievers.}  We further evaluate \model under different first-stage retrievers. As shown in Table~\ref{tab:first_stage}, \model consistently improves over all retrievers, and stronger retrievers yield better PRF and thus better reranking---demonstrating both robustness and the PRF-driven nature of the gains.

\paragraph{Generalization to another backbone.}
Applying Stage-II training to GTE-Qwen2-1.5B improves reranking while preserving its general embedding performance, showing that our recipe transfers beyond the Qwen3 family. Unlike the decoder-only Qwen3 models used in our main experiments, GTE-Qwen2 incorporates bidirectional attention, providing a structurally different test of the method. Detailed results are provided in Appendix~\ref{app:analysis}, Table~\ref{tab:gte}.

\section{Conclusion}

In this paper, we propose \model, a unified framework that enables a single text embedding model to perform both efficient retrieval and high-quality listwise reranking, by reformulating the listwise reranking prompt as a pseudo relevance feedback query. Extensive experiments demonstrate that \model can be an independent reranker and achieve state-of-the-art reranking performance on BEIR and strong results on BRIGHT, while significantly reducing inference latency compared to existing RankGPT-like listwise rerankers. Moreover, \model maintains competitive embedding capabilities on the MTEB benchmark.
Our work highlights the potential of single embedding models to serve as unified retrieval-reranking engines, offering a practical, efficient, and accurate alternative to complex multi-stage ranking systems.

\section*{Acknowledgments}

This research was sponsored by the Natural Science Foundation of China (62572475, 61902209, 62377044), Intelligent Social Governance Platform, Major Innovation Planning Interdisciplinary Platform for the ``Double-First Class'' Initiative, Renmin University of China, the Research Funds of Renmin University of China (22XNKJ15), and Beijing Nova Program.

\section*{Limitations}

This work focuses primarily on decoder-only LLM-based embedding models; broader validation on encoder-only architectures remains future work. The listwise prompt length grows with the number of candidate documents, which may pose challenges for very long documents or large candidate sets. Performance depends on first-stage feedback quality: controlled corruption is graceful on dense-relevance datasets, but can fall below the first stage when relevant evidence is extremely sparse. Training data construction relies on LLM-generated ranking labels without a separate human-label study, and their quality may vary across domains.

\section*{The Use of Large Language Models}

We used large language models (LLMs) only for portions of coding and language polishing. The authors reviewed all outputs and take full responsibility for the manuscript.

\bibliography{main}

\appendix

\section{Training Dataset Details}\label{app:dataset}

\paragraph{Dataset composition}

We mainly leverage the public portion of the E5 dataset~\citep{wang2023e5mistral}. Specifically, for the training at Stage I, we use the sampled version with around 1.5 million samples in total, which is constructed by~\cite{springer2025repetition} and is also used by LLM2Vec~\citep{behnamghaderllm2vec}. The mixture consists of ELI5, HotpotQA, FEVER, MIRACL, MS MARCO passage ranking and document ranking, NQ, NLI, SQuAD, TriviaQA, Quora Duplicate Questions, Mr.TyDi, DuReader, and T2Ranking. Each query in the datasets has only one positive and one negative.

As for the training at Stage II, since we need more negatives to meet the training objective, the E5 dataset cannot fully meet our requirements. Therefore, we used the dataset from BGE-M3~\citep{chen2024bgem3}, where each query contains multiple negatives. Specifically, we mainly used the retrieval dataset from the intersection of the E5 dataset and BGE-M3 dataset, including HotpotQA, MIRACL, MSMARCO passage, NQ, TriviaQA, DuReader, and T2Ranking. In addition, we have added two widely used Chinese retrieval datasets, cMedQAv2 and MMarco Chinese, which are included in the BGE-M3 dataset. Due to the length division of the BGE-M3 dataset, we only used the parts with document lengths less than 500. Meanwhile, we filtered queries containing fewer than 15 negative examples and further downsampled the dataset. In the end, we obtained a mixed dataset containing approximately 157k samples, with each instance containing one query, one negative, and fifteen negatives.

\begin{figure}[ht]
    \centering
    \includegraphics[width=\linewidth]{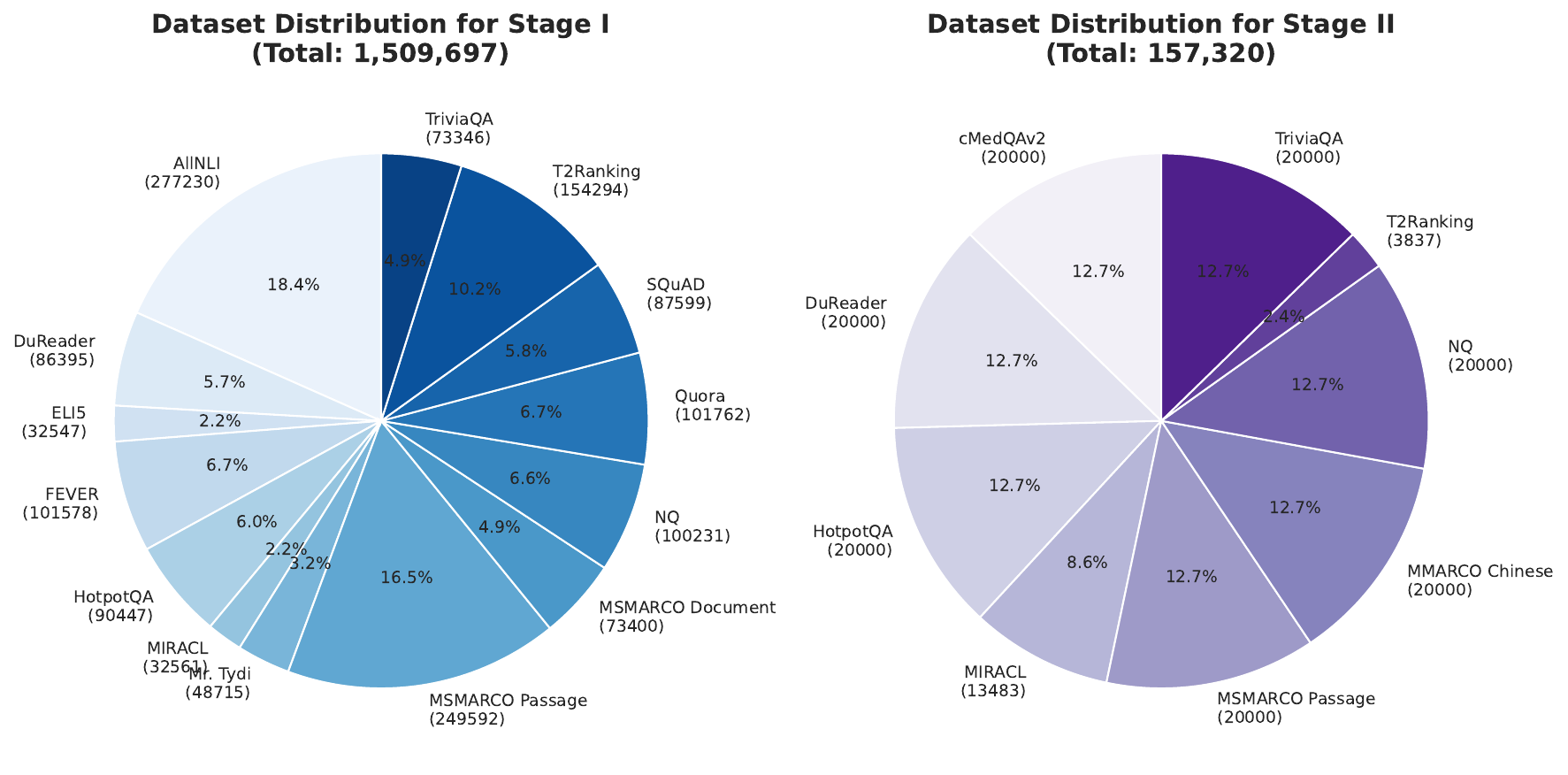}
    \caption{Dataset distribution for training.}
    \label{fig:dataset}
\end{figure}

\paragraph{Producing full ranking labels using Qwen3-32B}

We leverage Qwen3-32B (disabled thinking mode)~\citep{yang2025qwen3} to generate the full ranking labels for the Stage II training data. The process is similar to RankZephyr's~\citep {pradeep2023rankzephyr}. Specifically, we use the instruction in Table~\ref{inst:labeling} to have the model generate a ranking list in text form, and then parse the text. Then, we filter the results with the wrong output formations, which is only a very small portion of the entire dataset.

\begin{figure}[ht]
    \centering
    \includegraphics[width=0.8\linewidth]{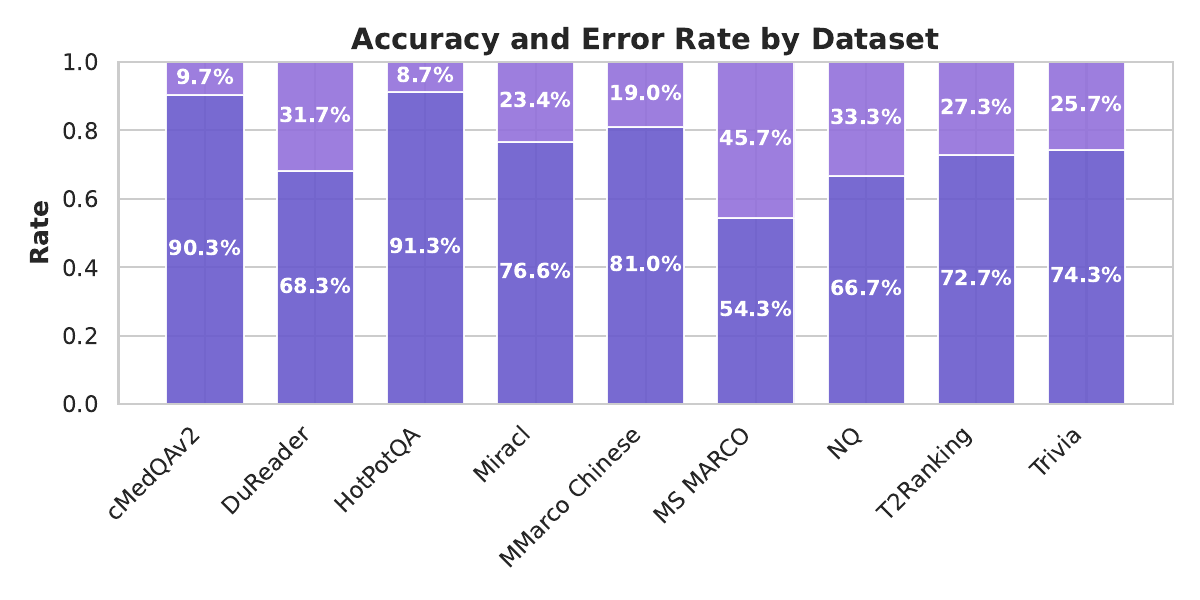}
    \caption{Accuracy for labeling the datasets.}
    \label{fig:dataset_acc}
\end{figure}

We measure label consistency as the frequency with which Qwen3-32B places the dataset's gold positive first (Figure~\ref{fig:dataset_acc}). Agreement is imperfect and only slightly above 50\% on MS MARCO, so we treat these labels as noisy supervision rather than ground truth. Nevertheless, the zero-shot teacher is a capable reranker: it reaches 53.37 average NDCG@10 on the BEIR subset (Table~\ref{tab:beir_full}) and 23.8 on BRIGHT (Table~\ref{tab:bright_bm25}), comparable to RankGPT-4o on BEIR. We did not conduct a separate human-label study at this scale. Prior work compares gold and reranker-generated labels without using LLM teachers~\citep{zhang2023rankwogpt}; higher-quality label construction remains future work.

\begin{table}[h]
\centering
\caption{Instruction for generating full ranking labels.}
\label{inst:labeling}
\setlength{\fboxsep}{8pt}
\fbox{\begin{minipage}{0.85\linewidth}\ttfamily\small
<|im\_start|>user

I will provide you with \{N\} passages, each indicated by a numerical identifier []. Rank the passages based on their relevance to the search query: \{query\}.

Documents:

[1] \{document 1\}

[2] \{document 2\}

...

[N] \{document N\}

Search Query: \{query\}

Rank the \{N\} passages above based on their relevance to the search query. All the passages should be included and listed using identifiers, in descending order of relevance. The output format should be [] > [] > ..., e.g., [4] > [2] > ..., Only respond with the ranking results, do not say anything else or explain. <|im\_end|>

<|im\_start|>assistant

<think>\textbackslash n\textbackslash n</think>\textbackslash n\textbackslash n
\end{minipage}}
\end{table}

\section{Implementation Details}\label{app:implementation}

In this section, we provide a detailed introduction to our training settings.

\paragraph{Stage I training} All models are trained with full parameters, DeepSpeed Zero3, brain floating point (BF16) quantization, and gradient checkpointing to optimize GPU memory consumption. We train on 8 NVIDIA A100 80G GPUs with an effective batch size of 512 for 1 epoch using a maximum sequence length of 512 tokens. We use a learning rate of $2 \times 10^{-5}$ and a linear learning rate warm-up for the first 300 steps. The instruction used for each dataset is adapted from~\cite{behnamghaderllm2vec}, which can be found in Table~\ref{tab:e5_insts}.

\begin{table*}[t]
\scriptsize
\centering
\caption{Instructions used for each of the E5 datasets in Stage I.}
\label{tab:e5_insts}
\begin{tabularx}{\linewidth}{lX}
\toprule
\textbf{Dataset} & \textbf{Instruction(s)} \\
\midrule
NLI & Given a premise, retrieve a hypothesis that is entailed by the premise \\
    & Retrieve semantically similar text \\
DuReader & Given a Chinese search query, retrieve web passages that answer the question \\
ELI5 & Provided a user question, retrieve the highest voted answers on Reddit ELI5 forum \\
FEVER & Given a claim, retrieve documents that support or refute the claim \\
HotpotQA & Given a multi-hop question, retrieve documents that can help answer the question \\
MIRACL & Given a question, retrieve Wikipedia passages that answer the question \\
MrTyDi & Given a question, retrieve Wikipedia passages that answer the question \\
MSMARCO Passage & Given a web search query, retrieve relevant passages that answer the query \\
MSMARCO Document & Given a web search query, retrieve relevant documents that answer the query \\
NQ & Given a question, retrieve Wikipedia passages that answer the question \\
QuoraDuplicates & Given a question, retrieve questions that are semantically equivalent to the given question \\
 & Find questions that have the same meaning as the input question \\
SQuAD & Retrieve Wikipedia passages that answer the question \\
T2Ranking & Given a Chinese search query, retrieve web passages that answer the question \\
TriviaQA & Retrieve Wikipedia passages that answer the question \\
\bottomrule
\end{tabularx}
\end{table*}

\paragraph{Stage II training}

For the training data, it is important to note that we do not use the full datasets introduced in Appendix~\ref{app:dataset} for training. Instead, for each dataset, we only sample at most 10,000 instances, leading to around 87k training instances actually. The instructions for the listwise prompt are listed in Table~\ref{tab:ranking_insts}.

\begin{table*}[t]
\scriptsize
\centering
\caption{Instructions in listwise prompts used for each of the datasets in Stage II.}
\label{tab:ranking_insts}
\begin{tabularx}{\linewidth}{lX}
\toprule
\textbf{Dataset} & \textbf{Instruction(s)} \\
\midrule
DuReader & Given a Chinese search query and some relevant documents, rerank the documents that answer the query \\
HotpotQA & Given a multi-hop question and some relevant documents, rerank the documents that answer the question \\
MIRACL & Given a question and some relevant Wikipedia documents, rerank the documents that answer the question \\
MSMARCO Passage & Given a web search query and some relevant documents, rerank the documents that answer the query \\
NQ & Given a question, retrieve Wikipedia passages that answer the question \\
T2Ranking & Given a Chinese search query and some relevant documents, rerank the documents that answer the query \\
TriviaQA & Given a question and some relevant Wikipedia documents, rerank the documents that answer the question \\
\bottomrule
\end{tabularx}
\end{table*}

We train all models on 8 NVIDIA A100 80G GPUs with an effective batch size of 128 for 1 epoch (each instance contains multiple documents). We use DeepSpeed Zero3, BF16, and gradient checkpointing to optimize GPU memory consumption. For documents, we use a maximum length of 1024 and in-batch negatives. Candidate documents are randomly shuffled when constructing Stage-II training instances, so relevant documents occur at varying positions; at inference time, the top-20 documents are supplied in first-stage rank order. We use a learning rate initialized at $5 \times 10^{-6}$ with a linear scheduler and a warmup ratio of 0.03.

\begin{table}[t]
\centering
\caption{Sensitivity to the Stage-II loss weight $\lambda$ using \model-0.6B trained on MS MARCO (NDCG@10).}
\label{tab:lambda_sensitivity}
\scriptsize
\begin{tabular}{@{}lcc@{}}
\toprule
$\lambda$ & \textbf{DL20} & \textbf{BEIR Avg.} \\
\midrule
0.5 & 69.76 & 49.69 \\
1.0 & 70.42 & 50.24 \\
2.0 (default) & \textbf{70.55} & \textbf{50.74} \\
3.0 & 70.43 & 50.67 \\
\bottomrule
\end{tabular}
\end{table}

\paragraph{Training RankQwen3}\label{app:rankqwen}

We fine-tune the Qwen3 model on the GPT-4 labeled listwise ranking dataset provided by~\cite{pradeep2023rankzephyr}. The dataset contains 40k samples, and we train the model for 1 epoch with a batch size of 16 per device, leading to an effective batch size of 64. For different sizes, we adjust the gradient accumulation steps to fit the batch size. We use DeepSpeed and BF16 mixed precision for acceleration. The learning rate is initialized at $5 \times 10^{-6}$ with a linear scheduler and a warmup ratio of 0.03. The training is performed on 4 NVIDIA A100 80G GPUs. We use LLM4Ranking Framework~\citep{liu2025llm4ranking} for training and evaluation.

\paragraph{Evaluation details}

We use the following instruction for the evaluation of all reranking tasks:
\begin{table}[h]
\centering
\setlength{\fboxsep}{8pt}
\fbox{\begin{minipage}{0.85\linewidth}\ttfamily\small
<|im\_start|>user

Given a web search query and some relevant documents, rerank the documents that answer the query:

Documents:

[1] \{document 1\}

[2] \{document 2\}

...

[N] \{document N\}

Search Query: \{query\}

<|im\_end|>

<|im\_start|>assistant

<think>\textbackslash n\textbackslash n</think>\textbackslash n\textbackslash n
\end{minipage}}
\end{table}

In fact, based on our experiments, different instructions have a very small impact on performance, at least not statistically significant. So this will not affect the experimental results of the paper.

\paragraph{Instructions used for evaluation of MTEB} When evaluating MTEB, we use the same instructions as~\cite{zhang2025qwen3embedding}. The list of instructions for each task is listed in Table~\ref{tab:mteb_instructions}.

\begin{table*}[t]
\centering
\scriptsize
\caption{Instructions used for evaluation on the MTEB benchmark.
``STS*'' refers to all the STS tasks.} \label{tab:mteb_instructions}
\begin{tabular}{{p{0.24\linewidth}p{0.7\linewidth}}}
\toprule
\textbf{Task Name} & \textbf{Instruction} \\
\midrule
AmazonCounterfactualClassif. & Classify a given Amazon customer review text as either counterfactual or not-counterfactual \\
AmazonPolarityClassification & Classify Amazon reviews into positive or negative sentiment  \\
AmazonReviewsClassification & Classify the given Amazon review into its appropriate rating category  \\
Banking77Classification & Given a online banking query, find the corresponding intents  \\
EmotionClassification &  Classify the emotion expressed in the given Twitter message into one of the six emotions: anger, fear, joy, love, sadness, and surprise  \\
ImdbClassification & Classify the sentiment expressed in the given movie review text from the IMDB dataset  \\
MassiveIntentClassification & Given a user utterance as query, find the user intents  \\
MassiveScenarioClassification & Given a user utterance as query, find the user scenarios  \\
MTOPDomainClassification & Classify the intent domain of the given utterance in task-oriented conversation  \\
MTOPIntentClassification & Classify the intent of the given utterance in task-oriented conversation  \\
ToxicConversationsClassif. & Classify the given comments as either toxic or not toxic  \\
TweetSentimentClassification & Classify the sentiment of a given tweet as either positive, negative, or neutral  \\
ArxivClusteringP2P & Identify the main and secondary category of Arxiv papers based on the titles and abstracts  \\
ArxivClusteringS2S & Identify the main and secondary category of Arxiv papers based on the titles  \\
BiorxivClusteringP2P & Identify the main category of Biorxiv papers based on the titles and abstracts  \\
BiorxivClusteringS2S & Identify the main category of Biorxiv papers based on the titles  \\
MedrxivClusteringP2P & Identify the main category of Medrxiv papers based on the titles and abstracts  \\
MedrxivClusteringS2S & Identify the main category of Medrxiv papers based on the titles  \\
RedditClustering & Identify the topic or theme of Reddit posts based on the titles  \\
RedditClusteringP2P & Identify the topic or theme of Reddit posts based on the titles and posts  \\
StackExchangeClustering & Identify the topic or theme of StackExchange posts based on the titles  \\
StackExchangeClusteringP2P & Identify the topic or theme of StackExchange posts based on the given paragraphs  \\
TwentyNewsgroupsClustering & Identify the topic or theme of the given news articles  \\
SprintDuplicateQuestions & Retrieve duplicate questions from Sprint forum  \\
TwitterSemEval2015 & Retrieve tweets that are semantically similar to the given tweet  \\
TwitterURLCorpus & Retrieve tweets that are semantically similar to the given tweet  \\
AskUbuntuDupQuestions & Retrieve duplicate questions from AskUbuntu forum  \\
MindSmallReranking & Retrieve relevant news articles based on user browsing history  \\
SciDocsRR & Given a title of a scientific paper, retrieve the titles of other relevant papers  \\
StackOverflowDupQuestions & Retrieve duplicate questions from StackOverflow forum  \\
ArguAna & Given a claim, find documents that refute the claim  \\
ClimateFEVER & Given a claim about climate change, retrieve documents that support or refute the claim  \\
CQADupstackRetrieval &  Given a question, retrieve detailed question descriptions from Stackexchange that are duplicates to the given question  \\
DBPedia & Given a query, retrieve relevant entity descriptions from DBPedia  \\
FEVER & Given a claim, retrieve documents that support or refute the claim  \\
FiQA2018 & Given a financial question, retrieve user replies that best answer the question  \\
HotpotQA & Given a multi-hop question, retrieve documents that can help answer the question  \\
MSMARCO & Given a web search query, retrieve relevant passages that answer the query  \\
NFCorpus & Given a question, retrieve relevant documents that best answer the question  \\
NQ & Given a question, retrieve Wikipedia passages that answer the question  \\
QuoraRetrieval & Given a question, retrieve questions that are semantically equivalent to the given question  \\
SCIDOCS & Given a scientific paper title, retrieve paper abstracts that are cited by the given paper  \\
SciFact & Given a scientific claim, retrieve documents that support or refute the claim  \\
Touche2020 & Given a question, retrieve detailed and persuasive arguments that answer the question  \\
TRECCOVID & Given a query on COVID-19, retrieve documents that answer the query  \\
STS* & Retrieve semantically similar text.  \\
SummEval & Given a news summary, retrieve other semantically similar summaries  \\
\bottomrule
\end{tabular}
\end{table*}

\section{Discussion of PRF-like Mechanism}\label{app:prf}

Pseudo-Relevance Feedback (PRF) has long been used in classical IR systems to refine the query representation using top-ranked documents, based on the assumption that these documents contain valuable signals that reflect the underlying relevance intent. Traditional PRF methods typically fall into two categories:
(1) text-based PRF, where top-retrieved documents are used to extract expansion terms or passages to enrich the query (e.g., Rocchio, RM3); and
(2) vector-based PRF, where the query embedding is updated by interpolating it with the embeddings of the top retrieved documents.
Both approaches require manually designed mechanisms---either for expansion term selection or for embedding fusion---and do not involve end-to-end optimization of how such feedback is interpreted.

In contrast, \model does not explicitly modify the query representation through heuristic expansion. Instead, it uses the top-N retrieved documents as implicit relevance feedback inputs and ``learns'' to interpret these documents jointly in a listwise context through supervision, capturing not only query-related relevance cues but also document--document relational signals. \model models the set-level interactions among these documents, which is essential for ranking.

For comparison, we implemented two classical PRF-style baselines using the model without listwise training (corresponding to "w/o Listwise in Stage II" in Table~\ref{tab:abl_training}), and applied (i) a text-based listwise prompt, and (ii) a vector-based Rocchio-style fusion of query and document embeddings.

As the results shown and discussed in Section~\ref{sec:analysis}, merely injecting PRF signals is ineffective or even harmful, and only a model like \model that learns supervised, ranking-aware feature transformations from candidate documents can achieve significant performance gains.

Interestingly, the behavior of \model exhibits characteristics similar to PRF: it benefits most from high-quality top-ranked documents and suffers when noisy or irrelevant documents (e.g., randomly sampled or tail documents) are used as feedback, as shown in Figure~\ref{fig:diff_docs} and Figure~\ref{fig:num_docs_score}. This supports the interpretation that \model uses top-ranked documents as relevance cues in a PRF-like manner. However, unlike classical PRF, \model does not rely on manually designed fusion mechanisms, but instead learns how to utilize these signals through end-to-end listwise supervision. This allows the model to determine how much each feedback document should contribute, what type of signal it provides, and how to incorporate it into the ranking-oriented embedding space, rather than merely enriching the query surface form or interpolating embedding vectors.

In summary, while \model is not a traditional PRF system, it operates under a learned PRF-like mechanism, where top-ranked documents provide weak relevance signals, but the ability to interpret, weight, and operationalize those signals is learned through listwise ranking supervision, rather than manually designed. We believe this inspires a new paradigm that bridges PRF intuition with modern embedding-based ranking.

\section{Analysis}\label{app:analysis}

\subsection{Influence of the Reranking Depth}

We conducted a comprehensive set of reranking experiments to evaluate the generalizability and robustness of our approach. Specifically, we varied both the reranking depth (Top-10, Top-20, Top-50, Top-100) and evaluated performance using three standard ranking metrics: NDCG@1, NDCG@5, and NDCG@10. The dataset we used is DL20 and BM25 is served as the first-stage retriever. This setup enables us to analyze not only the ability to correctly identify the single most relevant document, but also the overall relevance distribution across the top-ranked results.

\begin{figure*}[t]
    \centering
    \includegraphics[width=\textwidth]{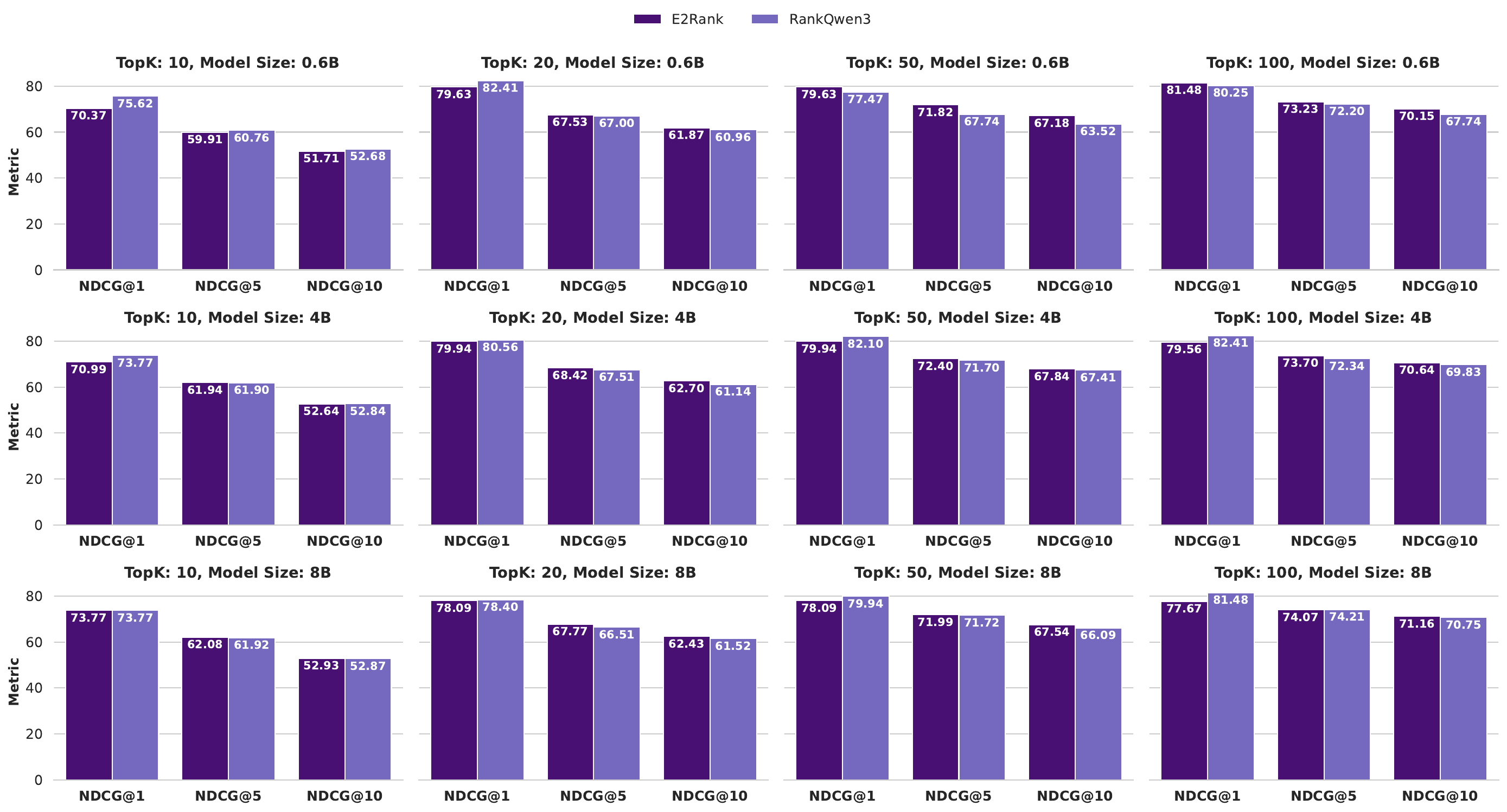}
    \caption{Results of different reranking settings on DL20.}
    \label{fig:topk_results}
\end{figure*}

Figure~\ref{fig:topk_results} presents a comparison between \model and RankQwen3 across all configurations. A clear trend emerges: \model consistently matches or outperforms RankQwen3 on NDCG@5 and NDCG@10, highlighting stronger listwise ranking capability and better top-K discrimination. RankQwen3 achieves higher NDCG@1 in several cases---particularly with smaller models or shallower reranking depths. We believe this is consistent with its generative design, which is effective when selecting the most relevant document at the first generative position. In contrast, \model optimizes the distribution of cosine similarity, making it inherently better at capturing global document relevance across the ranked list.

Furthermore, \model exhibits strong scalability, generalizing well across different ranking depths. Importantly, \model maintains competitive precision at rank 1 while delivering stronger performance at broader cutoffs, striking an effective balance between ranking quality and efficiency. These results collectively confirm the robustness, scalability, and practical applicability of \model in real-world reranking scenarios.

\subsection{Reranking Behavior}

We analyze the behavior of our rerankers along two dimensions: (i) their ability to refine the ranking in the head (top-ranked documents) and (ii) their ability to surface relevant documents from the tail (lower-ranked documents). We use a head cut-off $H$ (typically $H=20$) to distinguish the ``head'' (ranked $\leq H$) from the ``tail'' (ranked $> H$). For the analytical experiments in this section, we reranked the top 100 of BM25 on DL20.

\paragraph{Promotion and demotion analysis.}
To study the interaction between head and tail, we track which documents move into and out of the head when switching from the baseline to a reranker.
For each query $q$ and reranker, let $\pi(q,d)$ and $\pi_r(q,d)$ denote the rank positions assigned to $d$ by the first-stage retriever and the reranker, respectively (with $\pi(q,d) = \infty$ if $d$ is not retrieved).
We define:
\[
P_r(q) = \{ d \mid \pi(q,d) > H,\ \pi_r(q,d) \leq H \}
\]
\[
D_r(q) = \{ d \mid \pi(q,d) \leq H,\ \pi_r(q,d) > H \}
\]
corresponding to \emph{promoted} and \emph{demoted} documents, respectively.
For each reranker, we aggregate basic statistics over all $q$:
the total number of promoted or demoted documents, and the proportion of promoted or demoted documents are relevant (with relevance score $> 0$).
This reveals whether rerankers tend to replace less relevant head documents with more relevant tail documents.

\begin{table*}[t]
\centering
\scriptsize
\caption{Comparison of document promotion and demotion behavior across rerankers, showing how many documents are promoted from beyond rank 20 into the head, and the relevance quality of promoted versus demoted documents.}
\label{tab:promoted-demoted}
\setlength{\tabcolsep}{5pt}
\begin{tabular}{@{}l|c|c|cc@{}}
\toprule
Model          & NDCG@10 & \#Promoted (Demoted) & \% Rel. Promoted  & \% Rel. Demoted    \\
\midrule
RankQwen3-0.6B & 67.74   & 386            & 0.56  & 0.33  \\
RankQwen3-4B   & 69.83   & 405            & 0.61  & 0.31  \\
RankQwen3-8B   & 70.75   & 396            & 0.62  & 0.32  \\
\midrule
\model-0.6B    & 70.15   & 515            & 0.57  & 0.26  \\
\model-4B      & 70.64   & 527            & 0.56  & 0.27  \\
\model-8B      & 71.16   & 518            & 0.58  & 0.26  \\
\bottomrule
\end{tabular}
\end{table*}

Table~\ref{tab:promoted-demoted} compares how different rerankers modify the head of the ranking by promoting documents from the tail and demoting baseline head documents.
Across all models,  the fraction of promoted documents with $qrels>0$ is always higher than for demoted documents.
This indicates that all rerankers perform meaningful exchanges, generally replacing less relevant head documents with more relevant tail documents rather than perturbing the ranking arbitrarily.

Within the RankQwen3 family, larger models tend to make slightly higher-quality promotions: the fraction of relevant promoted documents increase from the 0.6B to the 8B, while the relevance of demoted documents decreases slightly.

Compared to RankQwen3, \model promote substantially more documents into the head, indicating a more aggressive reshaping of the top ranks.
The relevance profile of these exchanges is still favourable: promoted documents have higher relevance score, while demoted documents are markedly less relevant.
In other words, \model performs more head--tail swaps overall, and although each individual promotion is slightly less selective than for the strongest RankQwen3 model, the net effect is to clear out a larger number of low-quality head documents while maintaining a strong bias towards surfacing relevant items from the tail.

\paragraph{Decomposing head vs.\ tail contributions to NDCG@100.}
To disentangle improvements due to local reranking in the head from those due to promoting tail documents, we decompose the NDCG@100 gain of each reranker relative to the first-stage retriever.
We first construct a synthetic \emph{head-only} run that \emph{only} reranks the head, specifically, given the top-100 candidates of each query, rerank the documents ranked $\leq 20$ while keeping the other candidate set fixed. Then, we can define the NDCG@100 gain from only reranking the head, specifically, the difference of NDCG@100 between the head-only run and the original ranking, denoted as $g_{\text{within}}$. Similarly, we can define the NDCG@100 gain of the full run, i.e., the difference of NDCG@100 between the full run and the original ranking, denoted as $g_{\text{total}}$. The NDCG@100 gain from reranking the tail is defined as $g_{\text{tail}} = g_{\text{total}} - g_{\text{within}}$.

Intuitively, $g_{\text{within}}$ measures the gain attributable solely to reranking documents already in the head, while $g_{\text{tail}}$ captures the additional gain from promoting documents from the tail (or, more generally, changing the composition of the head).
We report mean and distributional statistics of these gains across queries to characterise each reranker as more ``head-refining'' or ``tail-mining''.

\begin{table*}[t]
\centering
\scriptsize
\caption{Decomposition of NDCG@100 gains into contributions from head reordering (within top-20) and tail promotions (beyond rank 20). The NDCG@100 of BM25 is 49.01. The numbers in parentheses represent the percentage improvement in NDCG@100 relative to the BM25 search results.}
\label{tab:gain_decomposition}
\begin{tabular}{@{}l|cccc@{}}
\toprule
Model & NDCG@100 & Mean Gain (Within-20) & Mean Gain (Tail) & Mean Gain (Total) \\
\midrule
RankQwen3-0.6B & 56.09 & 3.44 (7.0\%) & 3.64 (7.4\%) & 7.08 (14.4\%) \\
RankQwen3-4B   & 57.10 & 4.35 (8.9\%) & 3.74 (7.6\%) & 8.09 (16.5\%) \\
RankQwen3-8B   & 57.02 & 4.09 (8.3\%) & 3.92 (8.0\%) & 8.01 (16.3\%) \\
\midrule
E2Rank-0.6B    & 57.16 & 4.27 (8.7\%) & 3.88 (7.9\%) & 8.15 (16.6\%) \\
E2Rank-4B      & 56.95 & 4.38 (8.9\%) & 3.57 (7.3\%) & 7.94 (16.2\%) \\
E2Rank-8B      & 57.51 & 4.56 (9.3\%) & 3.94 (8.0\%) & 8.50 (17.3\%) \\
\bottomrule
\end{tabular}
\end{table*}

Table~\ref{tab:gain_decomposition} reports the decomposition of NDCG@100 improvements into gains from within-top-20 reranking and from promoting relevant documents from the tail. Across all models, the improvement in overall ranking quality is a combination of two complementary effects. The results consistently show that both contributions are substantial and of similar magnitude, confirming that all rerankers are capable of both head refinement and tail mining.

Within each model family, scaling generally improves total NDCG gain. Comparing model families, \model exhibits higher total gains than RankQwen3 at the same scale. Notably, \model tends to achieve slightly higher within-20 gains, indicating stronger head refinement ability, while maintaining competitive tail promotion capability. Meanwhile, RankQwen3 shows a more balanced pattern, with slightly lower head gains but comparable tail gains.

Overall, the results align with our previous observations: RankQwen3 scales toward more selective and precise head--tail exchanges, whereas \model not only makes more extensive replacements in the head but also converts those changes into slightly stronger global effectiveness improvements.

\subsection{Training From Existing Embedding Models}

In this work, we mainly chose LLMs (Qwen3 family) primarily because they offer stronger semantic understanding, richer contextual modeling, and longer context length than previous encoder-only models, which we believe is crucial for effectively capturing listwise interactions in the reranking stage. Additionally, it provided checkpoints of different sizes.

However, \model is not restricted to decoder-only models and is indeed generally applicable to different embedding backbones. To further validate generality, we have additionally applied our Stage-II training procedure to an existing encoder-based embedding model, GTE-Qwen2-1.5B (based on an LLM but integrated with bidirectional attention mechanisms, \cite{li2023towards}), without modifying its architecture.

\begin{table*}[t]
\centering
\caption{Results of training on GTE-Qwen2-1.5B.}
\label{tab:gte}
\scriptsize
\setlength{\tabcolsep}{2pt}
\begin{tabular}{ll|cc|ccccccccc|c}
\toprule
                                   & \textbf{Q Prompt} & \textbf{DL19} & \textbf{DL20} & \textbf{Covid} & \textbf{NFC.} & \textbf{Touche} & \textbf{DBPedia} & \textbf{SciFact} & \textbf{Signal} & \textbf{News} & \textbf{Robust} & \textbf{Avg.} & \textbf{MTEB(v2)} \\ \midrule
                                   & -            & 50.58         & 47.96         & 59.47               & 30.75             & 44.22               & 31.80            & 67.89            & 33.05             & 39.52              & 40.70             & 43.43             & -                      \\ \midrule
GTE-Qwen2-1.5B                     & Query-Only   & 68.07         & 61.97         & 82.75               & 37.94             & 36.25               & 41.86            & 77.43            & 31.10             & 50.29              & 54.86             & 51.56             & 67.20                  \\ \midrule
\multirow{2}{*}{+ \model Training} & Query-Only   & 71.31         & 66.20         & 83.52               & 38.48             & 35.38               & 43.56            & 77.16            & 32.68             & 48.65              & 57.63             & 52.13             & \multirow{2}{*}{67.19} \\
                                   & Listwise     & 69.98         & 68.97         & 81.16               & 39.49             & 46.17               & 41.81            & 75.47            & 34.37             & 53.55              & 56.51             & 53.57             &                        \\ \bottomrule
\end{tabular}
\end{table*}

The results in Table~\ref{tab:gte} show consistent improvements in both reranking benchmarks while maintaining the embedding performance, similar to those observed with the Qwen3 backbone. This supports our claim that the ranking objective and PRF-style listwise prompt are broadly effective and not tied to a particular model class.

\subsection{Robustness to Degraded PRF Windows}

We progressively replace relevant documents in the top-20 PRF window with non-relevant candidates while retaining the relevant documents in the top-100 scoring pool. Table~\ref{tab:prf_robustness} shows gradual degradation as feedback quality declines. Even when the window contains no relevant document, reranking remains above BM25 on all three dense-relevance datasets, indicating that feedback quality is a limitation but not the sole source of the gains.

\begin{table}[t]
\centering
\caption{Robustness to degrading the top-20 PRF window (NDCG@10, \model-0.6B). A fraction $p$ of relevant feedback documents is replaced by non-relevant candidates; relevant documents remain in the top-100 scoring pool.}
\label{tab:prf_robustness}
\scriptsize
\setlength{\tabcolsep}{4pt}
\begin{tabular}{@{}lccc@{}}
\toprule
\textbf{PRF window} & \textbf{DL19} & \textbf{DL20} & \textbf{COVID} \\
\midrule
BM25 (no reranking) & 50.58 & 47.96 & 59.47 \\
\midrule
$p=0$ (default) & 70.84 & 70.15 & 79.17 \\
$p=0.25$ & 69.74 & 69.08 & 77.52 \\
$p=0.50$ & 67.25 & 64.53 & 77.38 \\
$p=0.75$ & 65.07 & 58.69 & 71.38 \\
$p=1.00$ (no relevant feedback) & 55.27 & 51.55 & 63.27 \\
\bottomrule
\end{tabular}
\end{table}

\FloatBarrier
\section{Additional Experimental Results}\label{app:results}

\begin{table*}[t]
\centering
\scriptsize
\caption{Full results on BEIR. For reasoning rerankers, the results are borrowed from~\cite{liu2025reasonrank} and only contain 7 datasets, excluding Touche2020.}
\label{tab:beir_full}
\setlength{\tabcolsep}{4.7pt}
\begin{tabular}{l|cccccccc|cc}
\toprule
\textbf{} &
  \textbf{Covid} &
  \textbf{NFCorpus} &
  \textbf{Touche} &
  \textbf{DBPedia} &
  \textbf{SciFact} &
  \textbf{Signal} &
  \textbf{News} &
  \textbf{Robust} &
  \textbf{Avg. (7)} &
  \textbf{Avg. (8)} \\
\midrule
BM25                  & 59.47 & 30.75 & 44.22 & 31.80 & 67.89 & 33.05 & 39.52 & 40.70 & 43.31 & 43.43 \\
\midrule
\multicolumn{11}{l}{\textit{Previous Fine-tuned Pointwise Reranker}}             \\
\midrule
MonoBERT (340M)       & 70.01 & 36.88 & 31.75 & 41.87 & 71.36 & 31.44 & 44.62 & 49.35 & 49.36 & 47.16 \\
MonoT5 (3B)           & 79.80 & 37.30 & 32.20 & 48.30 & 76.30 & 32.50 & 44.80 & 58.50 & 53.93 & 51.21 \\
RankT5 (3B)           & 81.70 & 37.40 & 31.90 & 49.50 & 77.10 & 38.80 & 45.00 & 58.30 & 55.40 & 52.46 \\
\midrule
\multicolumn{11}{l}{\textit{Previous Fine-tuned Listwise Reranker}}             \\
\midrule
ListT5 (3B)           & 84.70 & 37.70 & 33.80 & 53.20 & 77.00 & 33.60 & 46.20 & 57.80 & 55.74 & 53.00 \\
RankVicuna            & 79.50 & 32.50 & 33.30 & 45.00 & 68.80 & 32.90 & 44.50 & 47.00 & 50.03 & 47.94 \\
FirstZephyr           & 82.76 & 31.13 & 26.81 & 42.48 & 73.93 & 32.51 & 48.53 & 51.30 & 51.81 & 48.68 \\
RankZephyr            & 83.20 & 37.60 & 32.40 & 44.50 & 74.90 & 31.50 & 52.50 & 54.30 & 54.07 & 51.36 \\
\midrule
\multicolumn{11}{l}{\textit{Zero-shot Listwise Reranker}}             \\
\midrule
RankGPT-4o            & 83.41      & 39.67    & 32.26      & 45.56   & 77.41   & 34.20    & 51.92     & 60.25    & 56.06        & 53.09        \\
RankGPT-4o-mini       & 80.03      & 38.73    & 30.91      & 44.54   & 73.14   & 33.64    & 50.91     & 57.41    & 54.06        & 51.16        \\
RankQwen3-14B         & 84.45 & 38.94 & 38.30 & 44.52 & 78.64 & 33.58 & 51.24 & 59.66 & 55.86 & 53.67 \\
RankQwen3-32B         & 83.48 & 39.22 & 37.13 & 45.00 & 78.22 & 32.12 & 51.08 & 60.74 & 55.69 & 53.37 \\
\midrule
\multicolumn{11}{l}{\textit{Reasoning Reranker}}             \\
\midrule
Rank-R1 (7B)          & 83.71 & 38.94 & -     & 42.27 & 72.16 & 33.08 & 50.60 & 54.46 & 53.60 & -     \\
Rank-R1 (14B)         & 84.63 & 38.58 & -     & 44.05 & 75.96 & 32.95 & 49.20 & 56.91 & 54.61 & -     \\
Rank1 (7B)            & 79.04 & 37.52 & -     & 35.79 & 73.32 & 25.41 & 47.67 & 57.11 & 50.84 & -     \\
Rearank (7B)          & 81.28 & 35.20 & -     & 45.23 & 75.02 & 36.00 & 51.88 & 57.49 & 54.59 & -     \\
ReasonRank (7B)       & 82.01 & 39.60 & -     & 46.03 & 75.55 & 31.36 & 50.50 & 55.40 & 54.35 & -     \\
\midrule
\multicolumn{11}{l}{\textit{Fine-tuned Listwise Reranker based on Qwen3}}             \\
\midrule
RankQwen3-0.6B        & 78.35 & 36.41 & 37.54 & 39.19 & 71.01 & 30.96 & 44.43 & 46.31 & 49.52 & 48.03 \\
RankQwen3-4B          & 83.91 & 39.88 & 32.66 & 43.91 & 76.37 & 32.15 & 50.81 & 59.36 & 55.20 & 52.38 \\
RankQwen3-8B          & 85.37 & 40.05 & 31.73 & 45.44 & 78.96 & 32.48 & 52.36 & 60.72 & 56.48 & 53.39 \\
\midrule
\multicolumn{11}{l}{\textit{Pointwise reranker finetund by RankNet loss based on Qwen3}}             \\
\midrule
Qwen3-0.6B (Pointwise)& 84.01 & 33.13 & 36.89 & 33.07 & 70.27 & 27.28 & 37.53 & 45.58 & 47.27 & 45.97 \\
Qwen3-4B (Pointwise)  & 80.40 & 31.58 & 29.92 & 40.84 & 72.09 & 25.98 & 47.56 & 56.60 & 50.72 & 48.12 \\
Qwen3-8B (Pointwise)  & 81.02 & 28.36 & 34.05 & 40.10 & 70.55 & 26.17 & 43.91 & 52.27 & 48.91 & 47.05 \\
\midrule
\multicolumn{11}{l}{\textit{Ours}}             \\
\midrule
\model-0.6B           & 79.17 & 38.60 & 41.91 & 41.96 & 73.43 & 35.26 & 52.75 & 53.67 & 53.55 & 52.09 \\
\model-4B             & 83.30 & 39.20 & 43.16 & 42.95 & 77.19 & 34.48 & 52.71 & 60.16 & 55.71 & 54.14 \\
\model-8B             & 84.09 & 39.08 & 42.06 & 43.44 & 77.49 & 34.01 & 54.25 & 60.34 & 56.10 & 54.35 \\
\bottomrule
\end{tabular}
\end{table*}

\paragraph{Full comparison on BEIR with baselines} We present all detailed results of baselines in Table~\ref{tab:beir_full}, which is an extended version of Table~\ref{tab:beir_all}. We report the results of reasoning-intensive rerankers, however, not all of them perform well on these general reranking tasks. In addition, we use the same training dataset with ~\model to train a cross-encoder style pointwise reranker, using the same RankNet loss. We believe that the reason why them do not perform so well is due to insufficient training data. This comparison between the pointwise model and \model also demonstrates the effectiveness of listwise reranking.

\paragraph{Full results for end-to-end retrieval} We present the detailed results of Table~\ref{tab:end2end} in Table~\ref{tab:beir_end2end} and Table~\ref{tab:bright_end2end}.

\begin{table*}[t]
\scriptsize
\centering
\caption{Full end-to-end ranking performance on BEIR.}
\label{tab:beir_end2end}
\begin{tabular}{@{}lr|ccccccccc@{}}
\toprule
 & &
  \textbf{Coivd} &
  \textbf{NFCorpus} &
  \textbf{Touche} &
  \textbf{DBPedia} &
  \textbf{SciFact} &
  \textbf{Signal} &
  \textbf{News} &
  \textbf{Robust} &
  \textbf{Avg.} \\ \midrule
\model-0.6b & Retrieval & 81.03 & 33.80    & 29.96  & 41.36   & 71.12   & 27.97  & 42.85 & 52.71    & 47.60 \\
            & + Rerank  & 83.33 & 37.62    & 30.87  & 43.68   & 72.95   & 27.94  & 50.03 & 58.89    & 50.66 \\
\midrule
\model-4b   & Retrieval & 81.84 & 38.64    & 27.95  & 47.75   & 78.94   & 27.90  & 49.56 & 64.29    & 52.11 \\
            & + Rerank  & 84.42 & 41.39    & 33.19  & 47.74   & 78.48   & 27.10  & 52.85 & 67.81    & 54.12 \\
\midrule
\model-8b   & Retrieval & 82.29 & 40.08    & 27.95  & 48.75   & 80.91   & 28.13  & 53.46 & 65.55    & 53.39 \\
            & + Rerank  & 86.61 & 42.33    & 34.86  & 48.20   & 78.99   & 26.31  & 53.75 & 69.58    & 55.08 \\ \bottomrule
\end{tabular}
\end{table*}

\begin{table*}[t]
\centering
\scriptsize
\setlength{\tabcolsep}{4pt}
\caption{Full end-to-end ranking performance on BRIGHT.}
\label{tab:bright_end2end}
\begin{tabular}{@{}ll|ccccccc|cc|ccc|c@{}}
\toprule
\multirow{2}{*}{\textbf{}} & \multirow{2}{*}{\textbf{}} &
  \multicolumn{7}{c|}{\textbf{StackExchange}} &
  \multicolumn{2}{c|}{\textbf{Coding}} &
  \multicolumn{3}{c|}{\textbf{Theorem-based}} &
  \multicolumn{1}{c}{\multirow{2}{*}{\textbf{Avg.}}} \\ \cmidrule(lr){3-14}
 & &
  \textbf{Bio.} &
  \textbf{Econ.} &
  \textbf{Earth.} &
  \textbf{Psy.} &
  \textbf{Rob.} &
  \textbf{Stack.} &
  \textbf{Sus.} &
  \textbf{Pony.} &
  \textbf{LC.} &
  \textbf{AoPS} &
  \textbf{TheoQ.} &
  \textbf{ThoT.} &
  \multicolumn{1}{l}{} \\
\midrule
\model-0.6b & Retrieval & 19.9 & 29.8 & 17.5 & 20.7 & 17.3 & 15.4 & 12.3 & 4.2 & 38.9 & 9.1  & 13.7 & 21.7 & 18.4 \\
            & + Rerank  & 27.1 & 37.4 & 23.1 & 31.0 & 22.4 & 19.3 & 20.0 & 3.7 & 38.8 & 8.9  & 17.9 & 21.5 & 22.6 \\
\midrule
\model-4b   & Retrieval & 35.4 & 42.6 & 23.7 & 34.4 & 24.5 & 22.2 & 22.4 & 7.2 & 43.0 & 11.3 & 33.5 & 34.1 & 27.8 \\
            & + Rerank  & 43.6 & 49.8 & 29.2 & 43.8 & 29.6 & 32.1 & 31.0 & 4.6 & 40.4 & 10.6 & 36.2 & 34.9 & 32.2 \\
\midrule
\model-8b   & Retrieval & 28.6 & 36.6 & 22.3 & 30.9 & 22.2 & 21.7 & 19.8 & 7.3 & 37.9 & 10.3 & 30.2 & 33.3 & 25.1 \\
            & + Rerank  & 39.9 & 46.6 & 28.9 & 41.7 & 28.3 & 29.7 & 34.4 & 6.1 & 37.4 & 9.1  & 33.6 & 36.7 & 31.0 \\
\bottomrule
\end{tabular}
\end{table*}

\paragraph{Full results for the ablation studies of training} We present the detailed results of Table~\ref{tab:abl_training} on Table~\ref{tab:beir_ablation}, Table~\ref{tab:bright_ablation}, and Table~\ref{tab:mtebv2_ablation}.

\begin{table*}[t]
\centering
\scriptsize
\caption{Full results of ablation study on BEIR.}
\label{tab:beir_ablation}
\begin{tabular}{l|ccccccccc}
\toprule
\textbf{} &
  \textbf{Coivd} &
  \textbf{NFCorpus} &
  \textbf{Touche} &
  \textbf{DBPedia} &
  \textbf{SciFact} &
  \textbf{Signal} &
  \textbf{News} &
  \textbf{Robust} &
  \textbf{Avg.} \\ \midrule
BM25                      & 59.47 & 30.75 & 44.22 & 31.80 & 67.89 & 33.05 & 39.52 & 40.70 & 43.43 \\ \midrule
\model-0.6B               & 79.17 & 38.60 & 41.91 & 41.96 & 73.43 & 35.26 & 52.75 & 53.67 & 52.09 \\ \midrule
w/o Stage I               & 79.22 & 38.13 & 40.98 & 40.99 & 73.18 & 33.35 & 51.74 & 53.01 & 51.33 \\
w/o InfoNCE in Stage II   & 79.48 & 39.02 & 40.87 & 42.27 & 74.27 & 34.42 & 52.44 & 54.59 & 52.17 \\ \midrule
w/ only Stage I           & 77.04 & 35.83 & 25.35 & 40.58 & 70.08 & 31.91 & 43.95 & 45.72 & 46.31 \\
w/o RankNet in Stage II   & 80.85 & 36.27 & 32.67 & 40.93 & 71.95 & 31.69 & 47.55 & 51.98 & 49.24 \\ \midrule
w/o Listwise in Stage II  & 81.47 & 36.79 & 33.33 & 41.41 & 72.44 & 31.93 & 48.83 & 53.24 & 49.93 \\ \bottomrule
\end{tabular}
\end{table*}

\begin{table*}[t]
\centering
\scriptsize
\setlength{\tabcolsep}{4.5pt}
\caption{Full results of ablation study on BRIGHT.}
\label{tab:bright_ablation}
\begin{tabular}{@{}l|ccccccc|cc|ccc|c@{}}
\toprule
\multirow{2}{*}{\textbf{}} &
  \multicolumn{7}{c|}{\textbf{StackExchange}} &
  \multicolumn{2}{c|}{\textbf{Coding}} &
  \multicolumn{3}{c|}{\textbf{Theorem-based}} &
  \multicolumn{1}{c}{\multirow{2}{*}{\textbf{Avg.}}} \\ \cmidrule(lr){2-13}
 &
  \textbf{Bio.} &
  \textbf{Econ.} &
  \textbf{Earth.} &
  \textbf{Psy.} &
  \textbf{Rob.} &
  \textbf{Stack.} &
  \textbf{Sus.} &
  \textbf{Pony.} &
  \textbf{LC.} &
  \textbf{AoPS} &
  \textbf{TheoQ.} &
  \textbf{ThoT.} &
  \multicolumn{1}{l}{} \\
\midrule
ReasonIR                  & 43.5 & 43.0 & 32.8 & 38.9 & 21.1 & 30.6 & 27.3 & 31.6 & 19.6 & 7.3  & 34.1 & 36.7 & 30.5 \\ 
\midrule
\model-0.6B               & 44.1 & 46.5 & 31.0 & 40.8 & 26.1 & 30.6 & 30.6 & 11.7 & 38.5 & 8.0  & 35.9 & 28.0 & 31.0 \\ \midrule
w/o Stage I               & 44.4 & 46.9 & 29.9 & 40.7 & 25.8 & 26.5 & 32.0 & 13.8 & 37.5 & 8.1  & 31.5 & 30.8 & 30.7 \\
w/o InfoNCE in Stage II   & 42.2 & 45.0 & 27.4 & 41.4 & 25.5 & 29.9 & 29.0 & 10.4 & 36.8 & 7.1  & 36.1 & 29.1 & 30.0 \\ \midrule
w/ only Stage I            & 11.1 & 16.7 & 13.0 & 13.3 & 14.8 & 11.0 & 10.3 & 10.0 & 37.4 & 9.0  & 14.4 & 22.7 & 15.3 \\
w/o RankNet in Stage II   & 25.1 & 36.2 & 22.0 & 26.2 & 19.8 & 20.1 & 16.2 & 8.0  & 40.2 & 9.7  & 20.2 & 25.0 & 22.4 \\ \midrule
w/o Listwise in Stage II  & 25.1 & 36.5 & 22.1 & 27.0 & 20.1 & 20.4 & 16.9 & 8.4  & 40.1 & 10.1 & 20.7 & 25.0 & 22.7 \\
\bottomrule
\end{tabular}
\end{table*}

\begin{table*}[t]
\centering
\scriptsize
\setlength{\tabcolsep}{3pt}
\caption{Detailed Results of ablation study on MTEB(eng, v2) Benchmark.}
\label{tab:mtebv2_ablation}
\begin{tabular}{@{}l|cccccc@{}}
\toprule
\textbf{Task} &
  \model-0.6B &
  \textbf{w/o Stage I} &
  \textbf{\begin{tabular}[c]{@{}c@{}}w/o InfoNCE\\ in Stage II\end{tabular}} &
  \textbf{w/ only Stage I} &
  \textbf{\begin{tabular}[c]{@{}c@{}}w/o RankNet\\ in Stage II\end{tabular}} &
  \textbf{\begin{tabular}[c]{@{}c@{}}w/o Listwise\\ in Stage II\end{tabular}} \\
\midrule
AmazonCounterfactualClassification     & 82.21 & 70.34 & 80.76 & 79.72 & 81.6  & 81.18 \\
ArXivHierarchicalClusteringP2P         & 58.55 & 57.87 & 57.09 & 57.38 & 58.52 & 58.13 \\
ArXivHierarchicalClusteringS2S         & 54.00 & 54.23 & 55.77 & 55.44 & 54.74 & 54.66 \\
ArguAna                                & 50.56 & 51.46 & 50.85 & 52.88 & 49.04 & 49.59 \\
AskUbuntuDupQuestions                  & 62.92 & 61.94 & 61.21 & 62.21 & 62.72 & 62.98 \\
BIOSSES                                & 85.22 & 85.35 & 85.13 & 84.68 & 85.54 & 85.64 \\
Banking77Classification                & 80.88 & 79.99 & 81.1  & 79.96 & 80.82 & 81.23 \\
BiorxivClusteringP2P.v2                & 39.28 & 38.63 & 38.27 & 38.50 & 38.82 & 40.17 \\
CQADupstackGamingRetrieval             & 55.35 & 53.32 & 55.36 & 56.19 & 56.33 & 57.02 \\
CQADupstackUnixRetrieval               & 39.31 & 38.3  & 39.27 & 41.53 & 40.42 & 40.6  \\
ClimateFEVERHardNegatives              & 30.80 & 28.91 & 28.03 & 26.37 & 30.53 & 30.91 \\
FEVERHardNegatives                     & 85.68 & 76.43 & 63.58 & 88.02 & 86.17 & 85.35 \\
FiQA2018                               & 40.84 & 36.79 & 34.74 & 38.12 & 40.88 & 40.91 \\
HotpotQAHardNegatives                  & 68.42 & 63.33 & 59.29 & 53.52 & 67.8  & 69.22 \\
ImdbClassification                     & 82.57 & 73.02 & 82.01 & 76.66 & 80.73 & 80.91 \\
MTOPDomainClassification               & 93.62 & 92.75 & 93.67 & 92.60 & 93.61 & 93.84 \\
MassiveIntentClassification            & 72.48 & 69.5  & 71.78 & 72.90 & 72.35 & 72.28 \\
MassiveScenarioClassification          & 74.71 & 72.92 & 74.9  & 74.58 & 74.57 & 74.82 \\
MedrxivClusteringP2P.v2                & 35.17 & 35.21 & 34.47 & 33.82 & 34.98 & 36.07 \\
MedrxivClusteringS2S.v2                & 31.19 & 30.82 & 32.17 & 32.61 & 31.18 & 32.04 \\
MindSmallReranking                     & 29.85 & 29.95 & 30.22 & 30.67 & 30.17 & 30.15 \\
SCIDOCS                                & 17.85 & 17.07 & 18.13 & 16.87 & 17.63 & 18.09 \\
SICK-R                                 & 79.89 & 70.59 & 80.63 & 79.69 & 79.81 & 79.9  \\
STS12                                  & 74.12 & 63.39 & 75.15 & 76.75 & 74.28 & 74.28 \\
STS13                                  & 84.19 & 80.41 & 84.93 & 84.07 & 83.83 & 84.83 \\
STS14                                  & 78.98 & 74.29 & 79.15 & 78.48 & 78.86 & 79.2  \\
STS15                                  & 86.25 & 82.54 & 86.57 & 85.99 & 86.41 & 86.68 \\
STS17                                  & 90.09 & 83.99 & 90.42 & 89.92 & 90.01 & 90.17 \\
STS22.v2                               & 65.60 & 65.08 & 65.53 & 60.30 & 62.9  & 63.78 \\
STSBenchmark                           & 84.75 & 79.05 & 85.46 & 84.39 & 84.58 & 84.88 \\
SprintDuplicateQuestions               & 93.49 & 94.73 & 92.67 & 91.15 & 93.78 & 93.82 \\
StackExchangeClustering.v2             & 53.12 & 53.99 & 52.16 & 56.38 & 52.82 & 54.59 \\
StackExchangeClusteringP2P.v2          & 38.95 & 39.1  & 38.94 & 38.91 & 39.19 & 39.77 \\
SummEvalSummarization.v2               & 31.66 & 28.75 & 31.63 & 31.55 & 31.12 & 31.02 \\
TRECCOVID                              & 81.03 & 78.78 & 67.55 & 70.48 & 81.11 & 82.01 \\
Touche2020Retrieval.v3                 & 58.46 & 56.42 & 51.36 & 53.79 & 59.77 & 58.44 \\
ToxicConversationsClassification       & 64.99 & 61.35 & 65.41 & 64.42 & 64.52 & 64.91 \\
TweetSentimentExtractionClassification & 66.23 & 62.33 & 66.38 & 66.04 & 66.08 & 65.81 \\
TwentyNewsgroupsClustering.v2          & 38.29 & 41.01 & 41.35 & 44.40 & 39.43 & 42.63 \\
TwitterSemEval2015                     & 72.13 & 65.49 & 69.93 & 70.68 & 72.08 & 71.41 \\
TwitterURLCorpus                       & 86.16 & 85.71 & 85.61 & 85.59 & 86.23 & 86.14 \\
\midrule
\textbf{Average}                       & 63.41 & 60.61 & 61.92 & 62.40 & 63.31 & 63.66 \\ \bottomrule
\end{tabular}
\end{table*}

\paragraph{Full results for the comparison with PRF baselines} We present the detailed results of Table~\ref{tab:prf} on Table~\ref{tab:beir_prf} and Table~\ref{tab:bright_prf}.

\begin{table*}[t]
\centering
\scriptsize
\caption{Full results of PRF baselines on BEIR.}
\label{tab:beir_prf}
\begin{tabular}{l|ccccccccc}
\toprule
\textbf{} &
  \textbf{Coivd} &
  \textbf{NFCorpus} &
  \textbf{Touche} &
  \textbf{DBPedia} &
  \textbf{SciFact} &
  \textbf{Signal} &
  \textbf{News} &
  \textbf{Robust} &
  \textbf{Avg.} \\ \midrule
BM25                      & 59.47 & 30.75 & 44.22 & 31.80 & 67.89 & 33.05 & 39.52 & 40.70 & 43.43 \\ \midrule
\model-0.6B               & 79.17 & 38.60 & 41.91 & 41.96 & 73.43 & 35.26 & 52.75 & 53.67 & 52.09 \\
\quad w/o Listwise Prompt & 81.35 & 36.27 & 33.50 & 41.09 & 71.88 & 31.70 & 47.15 & 52.70 & 49.46 \\ \midrule
w/o Listwise in Stage II  & 81.47 & 36.79 & 33.33 & 41.41 & 72.44 & 31.93 & 48.83 & 53.24 & 49.93 \\
\quad + text-based PRF    & 75.30 & 34.44 & 34.46 & 37.59 & 67.59 & 33.06 & 48.88 & 40.85 & 46.52 \\
\quad + vector-based PRF  & 78.73 & 37.08 & 38.06 & 40.66 & 70.50 & 30.61 & 48.12 & 49.84 & 49.20 \\
\bottomrule
\end{tabular}
\end{table*}

\begin{table*}[t]
\centering
\scriptsize
\setlength{\tabcolsep}{4.5pt}
\caption{Full results of PRF baselines on BRIGHT.}
\label{tab:bright_prf}
\begin{tabular}{@{}l|ccccccc|cc|ccc|c@{}}
\toprule
\multirow{2}{*}{\textbf{}} &
  \multicolumn{7}{c|}{\textbf{StackExchange}} &
  \multicolumn{2}{c|}{\textbf{Coding}} &
  \multicolumn{3}{c|}{\textbf{Theorem-based}} &
  \multicolumn{1}{c}{\multirow{2}{*}{\textbf{Avg.}}} \\ \cmidrule(lr){2-13}
 &
  \textbf{Bio.} &
  \textbf{Econ.} &
  \textbf{Earth.} &
  \textbf{Psy.} &
  \textbf{Rob.} &
  \textbf{Stack.} &
  \textbf{Sus.} &
  \textbf{Pony.} &
  \textbf{LC.} &
  \textbf{AoPS} &
  \textbf{TheoQ.} &
  \textbf{ThoT.} &
  \multicolumn{1}{l}{} \\
\midrule
ReasonIR                  & 43.5 & 43.0 & 32.8 & 38.9 & 21.1 & 30.6 & 27.3 & 31.6 & 19.6 & 7.3  & 34.1 & 36.7 & 30.5 \\ 
\midrule
\model-0.6B               & 44.1 & 46.5 & 31.0 & 40.8 & 26.1 & 30.6 & 30.6 & 11.7 & 38.5 & 8.0  & 35.9 & 28.0 & 31.0 \\
\quad w/o Listwise Prompt & 24.9 & 34.6 & 21.0 & 25.7 & 17.7 & 18.1 & 15.2 & 7.0  & 38.5 & 9.4  & 20.8 & 25.1 & 21.5 \\
\midrule
w/o Listwise in Stage II  & 25.1 & 36.5 & 22.1 & 27.0 & 20.1 & 20.4 & 16.9 & 8.4  & 40.1 & 10.1 & 20.7 & 25.0 & 22.7 \\
\quad + text-based PRF    & 49.3 & 49.4 & 31.2 & 22.0 & 22.0 & 30.6 & 27.0 & 26.5 & 33.9 & 7.0  & 34.2 & 22.5 & 29.6 \\
\quad + vector-based PRF  & 26.0 & 45.3 & 20.5 & 29.0 & 16.1 & 21.9 & 17.5 & 8.0  & 35.8 & 6.9  & 23.3 & 12.0 & 21.8 \\
\bottomrule
\end{tabular}
\end{table*}

\paragraph{Results using different first-stage retrieval models} Table~\ref{tab:first_stage} in Section~\ref{sec:analysis} reports results with the dense retrievers Contriver~\citep{izacard2021contriever}, BGE-base~\citep{xiao2023bge}, and Qwen3-Embedding-0.6B~\citep{zhang2025qwen3embedding}, as well as the neural sparse retriever SPLADE++ED~\citep{formal2022splade++ed}.

We report the reranking results on BRIGHT, using BM25 and the original query for first-stage retrieval, as presented in Table~\ref{tab:bright_bm25}. We also report the reranking results on BRIGHT, using BM25 and the GPT4 reasoning query for first-stage retrieval, as presented in Table~\ref{tab:bright_bm25cot}.

\begin{table*}[t]
\centering
\scriptsize
\setlength{\tabcolsep}{4.5pt}
\caption{Reranking results on BRIGHT. We use BM25 as the first-stage retriever and use original queries to obtain the top-100 candidates. The baseline results are mainly borrowed from~\cite{cai2025erank}. RankQwen3-14B (32B) are zero-shot, others are all fine-tuned.}
\label{tab:bright_bm25}
\begin{tabular}{@{}l|ccccccc|cc|ccc|c@{}}
\toprule
\multirow{2}{*}{\textbf{}} &
  \multicolumn{7}{c|}{\textbf{StackExchange}} &
  \multicolumn{2}{c|}{\textbf{Coding}} &
  \multicolumn{3}{c|}{\textbf{Theorem-based}} &
  \multicolumn{1}{c}{\multirow{2}{*}{\textbf{Avg.}}} \\ \cmidrule(lr){2-13}
 &
  \textbf{Bio.} &
  \textbf{Econ.} &
  \textbf{Earth.} &
  \textbf{Psy.} &
  \textbf{Rob.} &
  \textbf{Stack.} &
  \textbf{Sus.} &
  \textbf{Pony.} &
  \textbf{LC.} &
  \textbf{AoPS} &
  \textbf{TheoQ.} &
  \textbf{ThoT.} &
  \multicolumn{1}{l}{} \\
\midrule
BM25                 & 18.2 & 27.9   & 16.4  & 13.4 & 10.9 & 16.3   & 16.1 & 4.3   & 24.7 & 6.5  & 2.1   & 7.3    & 13.7 \\ 
\midrule
\multicolumn{14}{l}{\textit{Non-reasoning Listwise Reranker}}             \\
\midrule
RankZephyr           & 21.9 & 23.7   & 14.4  & 10.3 & 7.6  & 13.7   & 16.6 & 6.5   & 24.7 & 6.8  & 2.0   & 7.3    & 13.0 \\
RankQwen3-0.6B       & 21.2 & 32.3   & 17.4  & 20.8 & 14.7 & 14.9   & 18.8 & 6.0   & 26.6 & 6.4  & 4.5   & 8.7    & 16.0 \\
RankQwen3-4B         & 28.8 & 37.4   & 19.2  & 31.4 & 20.5 & 21.8   & 26.8 & 10.0  & 22.5 & 6.3  & 11.5  & 10.8   & 20.6 \\
RankQwen3-8B         & 29.7 & 40.2   & 21.0  & 31.0 & 23.3 & 23.2   & 27.0 & 10.1  & 16.9 & 6.5  & 12.0  & 11.6   & 21.1 \\
RankQwen3-14B        & 30.7 & 41.3   & 23.4  & 30.1 & 24.7 & 21.1   & 27.4 & 7.5   & 30.0 & 8.9  & 12.0  & 11.7   & 22.4 \\
RankQwen3-32B        & 31.9 & 45.5   & 23.8  & 33.2 & 25.6 & 22.5   & 30.8 & 7.5   & 29.7 & 10.9 & 11.7  & 13.0   & 23.8 \\
\midrule
\multicolumn{14}{l}{\textit{Reasoning-Intensive Reranker}}             \\
\midrule
Rank-R1-7B           & 26.0 & 28.5   & 17.2  & 24.2 & 19.1 & 10.4   & 24.2 & 4.3   & 19.8 & 4.3  & 10.9  & 8.3    & 16.4 \\
Rank1-7B             & 31.6 & 34.4   & 18.0  & 23.5 & 16.7 & 18.6   & 22.9 & 20.1  & 9.4  & 4.5  & 9.4   & 9.9    & 18.3 \\
Rearank-7B           & 23.4 & 27.4   & 18.5  & 24.2 & 17.4 & 16.3   & 25.1 & 8.0   & 27.0 & 7.4  & 9.5   & 7.9    & 17.7 \\
JudgeRank-8B         & 28.7 & 32.2   & 20.9  & 24.6 & 16.5 & 18.3   & 20.6 & 11.7  & 7.1  & 4.7  & 8.4   & 10.0   & 17.0 \\
ERank-4B             & 30.4 & 42.5   & 21.5  & 27.7 & 22.4 & 22.9   & 24.0 & 31.6  & 14.6 & 11.0 & 12.1  & 11.4   & 22.7 \\
ERank-14B            & 31.2 & 43.6   & 25.8  & 27.8 & 23.1 & 23.9   & 24.6 & 29.8  & 16.8 & 8.6  & 10.5  & 11.9   & 23.1 \\
\midrule
\multicolumn{14}{l}{\textit{Ours}}             \\
\midrule
\model-0.6B          & 27.1 & 41.7   & 20.7  & 24.3 & 19.8 & 22.1   & 19.6 & 4.8   & 32.7 & 10.7 & 8.6   & 9.9    & 20.2 \\
\model-4B            & 27.8 & 45.6   & 23.9  & 27.6 & 21.4 & 25.0   & 24.3 & 5.0   & 34.8 & 12.3 & 9.3   & 10.8   & 22.3 \\
\model-8B            & 28.7 & 45.2   & 24.4  & 27.2 & 22.9 & 25.2   & 25.6 & 6.6   & 32.5 & 11.8 & 9.3   & 10.7   & 22.5 \\
\bottomrule
\end{tabular}
\end{table*}

\begin{table*}[t]
\centering
\scriptsize
\setlength{\tabcolsep}{4.5pt}
\caption{Reranking results on BRIGHT. We use BM25 as the first-stage retriever and use GPT4 reasoning-queries to obtain the top-100 candidates. The baseline results are mainly borrowed from~\cite{cai2025erank}.}
\label{tab:bright_bm25cot}
\begin{tabular}{@{}l|ccccccc|cc|ccc|c@{}}
\toprule
\multirow{2}{*}{\textbf{}} &
  \multicolumn{7}{c|}{\textbf{StackExchange}} &
  \multicolumn{2}{c|}{\textbf{Coding}} &
  \multicolumn{3}{c|}{\textbf{Theorem-based}} &
  \multicolumn{1}{c}{\multirow{2}{*}{\textbf{Avg.}}} \\ \cmidrule(lr){2-13}
 &
  \textbf{Bio.} &
  \textbf{Econ.} &
  \textbf{Earth.} &
  \textbf{Psy.} &
  \textbf{Rob.} &
  \textbf{Stack.} &
  \textbf{Sus.} &
  \textbf{Pony.} &
  \textbf{LC.} &
  \textbf{AoPS} &
  \textbf{TheoQ.} &
  \textbf{ThoT.} &
  \multicolumn{1}{l}{} \\
\midrule
BM25              & 18.2 & 27.9   & 16.4  & 13.4 & 10.9 & 16.3   & 16.1 & 4.3   & 24.7 & 6.5  & 2.1   & 7.3    & 13.7 \\ 
\midrule
\multicolumn{14}{l}{\textit{Non-reasoning Listwise Reranker}}             \\
\midrule
RankQwen3-0.6B    & 26.8   & 50.5 & 46.5   & 23.6  & 37.7 & 21.5 & 27.7   & 26.6 & 19.7  & 21.3 & 4.4  & 22.6  & 20.1   \\
RankQwen3-4B      & 31.7   & 49.6 & 47.6   & 25.8  & 44.5 & 27.3 & 30.9   & 37.5 & 31.2  & 20.8 & 6.8  & 32.7  & 25.6   \\
RankQwen3-8B      & 31.6   & 50.8 & 47.0   & 26.8  & 45.4 & 28.0 & 30.5   & 38.1 & 27.4  & 20.0 & 6.4  & 32.4  & 26.8   \\
\midrule
\multicolumn{14}{l}{\textit{Reasoning-Intensive Reranker}}             \\
\midrule
Rank-R1-7B        & 23.9   & 38.2 & 29.4   & 23.4  & 33   & 24.9 & 14.9   & 33.2 & 18.2  & 16.1 & 3.8  & 16.6  & 34.8   \\
Rank1-7B          & 25.5   & 45.8 & 37     & 22.2  & 31.7 & 20.6 & 23     & 34.2 & 15.7  & 19.8 & 1.3  & 19.8  & 34.7   \\
Rearank-7B        & 29.1   & 42   & 37.5   & 26.4  & 39.1 & 25   & 25.1   & 32.6 & 26.2  & 29.2 & 5.9  & 28    & 32.2   \\
JudgeRank-8B      & 24.4   & 41.4 & 34.7   & 26.2  & 36   & 24   & 27.6   & 26.1 & 10.2  & 14.2 & 3.4  & 20.3  & 28.9   \\
ERank-4B          & 32.9   & 48.2 & 46.7   & 30    & 43.1 & 28.4 & 31.5   & 38.1 & 28.5  & 23.5 & 10.4 & 26.9  & 39.0   \\
ERank-14B         & 33.5   & 51.4 & 48.6   & 30.8  & 41.3 & 26.7 & 35.6   & 39.1 & 27.3  & 26.4 & 10.9 & 25.7  & 37.9   \\
\midrule
\multicolumn{14}{l}{\textit{Ours}}             \\
\midrule
\model-0.6B         & 29.6   & 47.6 & 52.2   & 27.4  & 42.0 & 26.8 & 31.5   & 29.5 & 19.5  & 31.8 & 7.5  & 19.2  & 20.5   \\
\model-4B           & 32.0   & 49.6 & 51.1   & 31.6  & 43.8 & 29.0 & 33.7   & 33.2 & 16.4  & 31.9 & 7.6  & 33.2  & 22.9   \\
\model-8B           & 32.6   & 51.7 & 51.4   & 32.1  & 45.6 & 27.3 & 35.5   & 34.5 & 16.2  & 31.1 & 8.3  & 32.7  & 25.2   \\
\bottomrule
\end{tabular}
\end{table*}

\end{document}